\documentclass[]{fairmeta}

\title{Personal-Agent Mediated Recommendation with Cross-Platform User History}

\author[1,*]{Yu Xia}
\author[2]{Jiangfan Zhang}
\author[2]{Jun Xiao}
\author[1]{Julian McAuley}
\author[2]{Xiangjun Fan}

\affiliation[1]{University of California San Diego}
\affiliation[2]{Meta AI}

\contribution[*]{Work done during Yu Xia's internship at Meta, mentored by Jiangfan Zhang.}

\abstract{Modern recommendation is shifting from platform-centric personalization toward user-governed personalization, where a personal LLM agent can act on the user's behalf across services. We formalize this emerging paradigm as \textbf{Personal-Agent Mediated Recommendation}: a platform recommender ranks a candidate set using platform-local information, and a personal agent uses user-authorized cross-platform history to mediate the resulting ranking and produce the final top-$K$ slate. Such mediation is nontrivial: the platform ranking can encode strong population evidence that the personal agent cannot observe, so effective mediation must therefore balance beneficial rescues against harmful overrides. 
To study this trade-off, we introduce \textbf{{MediateRec}}, a benchmark that includes scalable proxy cross-platform environments and a real cross-platform test under a controlled platform--agent information boundary. To train the agent to use cross-platform history effectively, we further propose \textbf{Personal Attribution Mediation Optimization (PAMO)}, which counterfactually masks that history to estimate personal mediation support and reallocates rank-aware advantage mass under a platform-relative value floor. We theoretically prove that PAMO preserves cutoff-level advantage mass and is locally optimal among first-order reallocations that preserve this mass without lowering average platform-relative value. Experiments on {MediateRec} show that personal-agent mediation enables meaningful platform corrections, yet even strong proprietary LLMs introduce non-negligible harmful overrides. PAMO consistently improves over matched outcome-only RL across seen and unseen target platforms and on the real cross-platform test, while achieving a better rescue--harm balance.}

\date{\today}

\contact{
  \email{yux078@ucsd.edu},
  \email{jiangfanzhang@meta.com},
  \email{junxiao@meta.com},\\
  \email{jmcauley@ucsd.edu},
  \email{maxfan@meta.com}
}

\usepackage{wrapfig}
\usepackage{tabularx}
\usepackage{array}
\usepackage{enumitem}
\usepackage{booktabs}
\usepackage{graphicx}
\usepackage[most]{tcolorbox}
\usepackage{amsthm}
\usepackage{amsfonts}
\usepackage{amssymb}
\newtheoremstyle{pamostyle}
  {4pt}          
  {4pt}          
  {\itshape}     
  {0pt}          
  {\bfseries}    
  {.}            
  {0.5em}        
  {}             

\theoremstyle{pamostyle}
\newtheorem{proposition}{Proposition}
\newtheorem{theorem}{Theorem}

\begin{document}

\maketitle

\section{Introduction}

Personalization in recommender systems has traditionally been confined to individual platforms, each modeling users from the interactions observed within its own service. The emergence of user-authorized personal LLM agents such as Muse \citep{meta2026muse} points to a more user-governed architecture, in which an agent can maintain user history across services and act on the user's behalf \citep{zhang2026usercentric,lin2026usergoverned,liu2026governable,sun2026autonomousagents}. Platforms can exploit platform-local interactions and population-level collaborative patterns that the personal agent cannot directly observe \citep{covington2016youtube,kang2018sasrec}, while personal agents can use cross-platform history unavailable to any single service.

\begin{figure}[t]
\centering
\includegraphics[width=\textwidth]{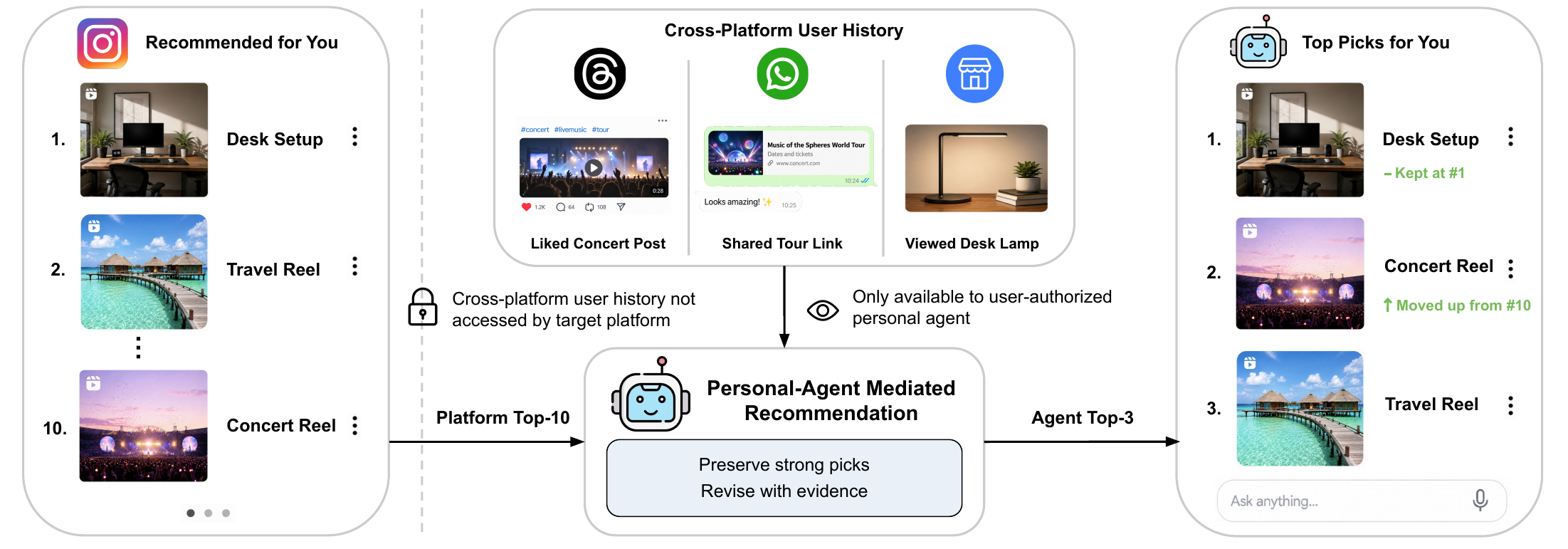}
\vspace{-0.5em}
\caption{Personal-Agent Mediated Recommendation. The platform proposal
summarizes platform-local population evidence, while the personal agent
additionally uses cross-platform history to selectively revise or preserve
the ranking.}
\label{fig:overview}
\end{figure}

Recent agentic recommender systems equip LLM agents with planning, memory, tool use, interaction, and multi-agent coordination \citep{huang2025agenticrec,lin2026agenticroadmap,shang2025agentrecbench,xia2026macf}. These systems mainly position the agent as a platform-side recommender or as a component inside the platform pipeline. They do not directly address a user-side personal agent that receives an existing platform proposal and uses cross-platform history to revise it. This leaves a concrete question: when does cross-platform history justify changing a strong platform ranking, and when should the agent defer to it?

We formalize this setting as \textbf{Personal-Agent Mediated Recommendation}. A platform recommender first ranks a candidate set using platform-local information. A personal agent then receives the resulting platform proposal together with the user's cross-platform history and produces the final top-$K$ slate. The platform ranking is a strong default: it can encode collaborative patterns across users that the personal agent cannot observes. The agent may \emph{rescue} a target that the platform misses, but it may also introduce a \emph{harmful override} by dropping a target that platform ranks correctly. Effective mediation therefore requires correcting platform misses without weakening platform decisions. This rescue--harm trade-off distinguishes mediation from standalone reranking and makes selective use of cross-platform history its central learning problem.

Evaluating this task requires paired information views of the same recommendation episode: a platform ranking generated from platform-local information and cross-platform user history exposed only to the personal agent. Existing agentic recommendation benchmarks \citep{shang2025agentrecbench,narasimhan2026taurec} mainly evaluate how well platform-side agents use within-platform information rather than evaluating a personal agent mediating a given platform ranking. 
To this end, we introduce \textbf{{MediateRec}}, a benchmark that constructs a controlled platform--agent information boundary: the platform proposal is generated from platform-local information, while cross-platform history is available only to the personal agent. {MediateRec} includes scalable proxy cross-platform environments built from Amazon Reviews \citep{hou2024blair} with a real cross-platform external test built from linked activity across independent services \citep{ballou2025openplay}. 
Results on {MediateRec} show that access to cross-platform history alone does not ensure effective mediation: strong proprietary LLMs are able to make meaningful platform corrections, yet still introduce non-negligible harmful overrides.

Recent methods commonly train recommendation agents with ranking or task rewards \citep{lin2025recr1,huang2026rerec,zhu2025rankgrpo,liu2025onerecthink}. In personal-agent mediated recommendation, however, final outcomes do not distinguish equally rewarded responses by how strongly they depend on the user's cross-platform history. A successful correction may genuinely rely on that history or may arise from generic reranking. To train the personal agent to use cross-platform history more effectively, we propose \textbf{Personal Attribution Mediation Optimization (PAMO)}. PAMO samples responses with full history and rescores the same visible rationales after masking cross-platform history. The likelihood contrast defines a \emph{personal mediation support} score measuring each rationale’s dependence on cross-platform history. Recommendation outcomes determine the sign and total amount of advantage, personal mediation support guides how that advantage is allocated among responses that disagree with the platform, and a value floor keeps the reallocation aligned with platform-relative recommendation utility.

We prove that PAMO preserves the positive and negative advantage mass induced by ranking outcomes and is locally optimal among first-order reallocations that preserve this mass without lowering average platform-relative value. Empirically, PAMO improves over matched outcome-only RL with better recommendation accuracy and rescue--harm balance across seen and unseen target platforms and on the real cross-platform test.
Our main contributions are threefold:
\begin{itemize}[left=0pt]
\item We formalize \textbf{Personal-Agent Mediated Recommendation}, where a personal agent uses cross-platform history to mediate a platform proposal and produce the final recommendation, and characterize its platform-relative rescues and harmful overrides.
\item We introduce \textbf{{MediateRec}}\footnote{The MediateRec benchmark will be released upon internal approval to facilitate future research.}, which includes scalable proxy cross-platform environments and a real cross-platform external test under a controlled platform--agent information boundary.
\item We propose \textbf{Personal Attribution Mediation Optimization (PAMO)}, which estimates personal mediation support through counterfactual masking and performs value-preserving advantage reallocation. We establish theoretically its local optimality among first-order reallocations that preserve this mass without decreasing platform-relative value.
\end{itemize}

\section{Related Work}

\subsection{Agentic Recommender Systems}

Agentic recommender systems augment platform-side recommendation with LLM reasoning, tool use, memory, interaction, and multi-agent coordination \citep{huang2025agenticrec,lin2026agenticroadmap}. Tool-augmented agents acquire user, item, and collaborative information during recommendation \citep{huang2023interecagent,wang2023recmind,zhang2026recthinker}, while multi-agent methods coordinate user- and item-oriented reasoning \citep{zhang2023agentcf,xia2026macf}. Memory-based systems maintain evolving user states through collaborative or hierarchical memory structures \citep{chen2026memrec,shen2026mars}. 
These methods strengthen the platform-side recommender, whose platform proposal forms the input to our mediation setting.
Recommendation agents are commonly adapted through memory and skill evolution, supervised trajectory tuning, and reinforcement learning from recommendation feedback. MemRec \citep{chen2026memrec} and MARS \citep{shen2026mars} update what the agent remembers about the user, while SAGER \citep{tao2026sager} evolves user-specific reasoning skills. Post-training methods \citep{zhang2026recthinker,lin2025recr1,huang2026rerec,zhu2025rankgrpo,liu2025onerecthink} teach recommendation reasoning and optimize ranking outcomes through task-specific rewards or finer-grained credit assignment. Our proposed PAMO instead targets personal mediation over an existing platform ranking, combining recommendation outcomes with support contributed by cross-platform history.

\subsection{User-Governed Personalization and Agents}

Personal agents adapt planning and actions to individual users through profile modeling, persistent memory, and interaction history \citep{xu2026personalizedagents}. Representative systems connect user-specific memory to tool use, mobile interaction, and GUI actions \citep{zhang2026personaagent,wang2026meagent,lyu2026personalalign}, while PersonaLens \citep{zhao2025personalens} and Persona2Web \citep{kim2026persona2web} evaluate preference inference from prior user records. Recent position papers \citep{zhang2026usercentric,lin2026usergoverned,liu2026governable,sun2026autonomousagents} place these capabilities within a user-governed architecture where personal agents maintain context across services and mediate interactions with platforms, which our formulation instantiates for recommendation.
A similar user--agent--platform system, iAgent
\citep{xu2025iagent}, centers personalization on explicit user instructions, tools, and feedback memory. Although it is framed as reranking an initial platform list, it constructs each candidate list by randomly sampling one target item and nine negatives, rather than mediating a scored platform ranking. Our setting instead starts from a frozen, strong platform proposal and studies when cross-platform history should revise or preserve it.
A concurrent work, ClawRec \citep{wu2026clawrec}, directly constructs a unified cross-source recommendation slate from multi-source context, whereas we train a personal agent to mediate a strong platform proposal using cross-platform history. Our {MediateRec} benchmark is built from public interaction traces and includes a test-only external evaluation on real linked services.
Cross-domain recommendation typically combines multi-source or multi-domain behavior within a centralized recommender \citep{li2021recguru,hou2022unisrec,ju2025autocdsr}. Our setting instead holds the target platform recommender unchanged and introduces cross-platform history only at user-side personal-agent mediation. 
In {MediateRec}, we use cross-domain records in Amazon Reviews \citep{hou2024blair} to construct scalable proxy platform boundaries, while OpenPlay \citep{ballou2025openplay} evaluates the same information boundary across real platform services.

\section{Personal-Agent Mediated Recommendation}\label{sec:formulation}

\subsection{Task Formulation}

For each recommendation episode, the platform returns a proposal
\begin{equation}
B=(\mathcal C,\rho_P,M),
\qquad
P=\operatorname{Top}_K(\rho_P),
\label{eq:platform_output}
\end{equation}
where $\mathcal C$ is the candidate set, $\rho_P$ is the complete platform ranking over $\mathcal C$, $M$ contains candidate metadata, and $P$ is the platform's default top-$K$ slate.
Let
\begin{equation}
    H=(H^{\mathrm{within}},H^{\mathrm{cross}})
\end{equation}
denote the user-authorized history available to the personal agent, where $H^{\mathrm{within}}$ contains target-platform records and $H^{\mathrm{cross}}$ contains cross-platform records unavailable to the target platform. Given the platform proposal and history, the agent generates a visible rationale $R$ followed by a ranked final slate:
\begin{equation}
(R,S)\sim\pi_\theta(\cdot\mid B,H),
\qquad
S\in\Pi_K(\mathcal C),
\label{eq:mediation_policy}
\end{equation}
where $\Pi_K(\mathcal C)$ is the set of ordered length-$K$ selections from $\mathcal C$.
The platform ranking summarizes platform-local evidence, including population-level collaborative patterns that the personal agent cannot directly observe. Conversely, the platform does not observe $H^{\mathrm{cross}}$. The agent receives the platform proposal but not the platform's scores or internal state, making the information asymmetry two-sided. Since returning the platform slate $P$ is valid, the task is selective mediation rather than ranking from scratch.

\subsection{Mediation Objective}

Let $Y$ denote the held-out target item. For any slate $X$, set $\operatorname{rank}_X(Y)=\infty$ when $Y\notin X$. We consider top-$K$ inclusion and single-target NDCG@$K$:
\begin{equation}
\begin{aligned}
U_{\mathrm{HR}}(X,Y)
&=
\mathbf 1\{Y\in X\},
\\
U_{\mathrm{NDCG}}(X,Y)
&=
\frac{
\mathbf 1\{Y\in X\}
}{
\log_2(\operatorname{rank}_X(Y)+1)
}.
\end{aligned}
\label{eq:ranking_utilities}
\end{equation}
Let $U$ denote the chosen recommendation utility. Training maximizes $\mathbb E_{S\sim\pi_\theta(\cdot\mid B,H)}[U(S,Y)]$. To characterize the agent's effect relative to the platform, we define the episode-level mediation value
\begin{equation}
\Delta(S;P,Y)
=
U(S,Y)-U(P,Y).
\label{eq:mediation_value}
\end{equation}
Because $P$ is fixed with respect to $\pi_\theta$, maximizing $\mathbb E[\Delta(S;P,Y)]$ is equivalent to maximizing $\mathbb E[U(S,Y)]$. The platform-relative form distinguishes beneficial corrections from harmful changes to the platform slate.

For top-$K$ inclusion, mediation has four outcomes: a \emph{preserved hit} when $Y\in P\cap S$, a \emph{rescue} when $Y\notin P$ but $Y\in S$, a \emph{harmful override} when $Y\in P$ but $Y\notin S$, and a \emph{mutual miss} otherwise. The resulting platform-relative improvement is
\begin{equation}
\Delta\mathrm{HR}@K
=
\Pr(\mathrm{rescue})
-
\Pr(\mathrm{harmful\ override}).
\label{eq:rescue_harm_decomposition}
\end{equation}
For NDCG@$K$, the mediation value in Eq.~\eqref{eq:mediation_value} also captures promotions and demotions of the target item. Effective mediation therefore requires making high-value corrections while avoiding degradation of strong platform decisions.

\section{{MediateRec} Benchmark}
\label{sec:benchmark}

We introduce
{MediateRec} to support training and evaluation under the platform--agent information boundary defined above. The benchmark includes four proxy cross-platform environments built from Amazon Reviews for scalable training and evaluation and a real cross-platform external test built from OpenPlay.
The platform is constructed using only within-platform information, whereas the personal agent additionally receives cross-platform history. 

\subsection{Benchmark Design}

\subsubsection{\textbf{Amazon Proxy Platforms}}

Public interaction data rarely link the same user across independent services because platform records are siloed and privacy-sensitive \citep{lin2026usergoverned}. We therefore use Amazon product categories as controlled proxy platforms. We select original \emph{Movies \& TV}, \emph{Toys \& Games}, \emph{Grocery \& Gourmet Food}, and \emph{Beauty \& Personal Care} categories as the target platforms for Movie, Toy, Grocery, and Beauty, respectively. Earlier interactions on the target platform form \emph{within-platform history}, and earlier interactions on all other Amazon categories of the same user form \emph{cross-platform history}. Although Amazon categories are not independent services, this construction preserves the information asymmetry of interest: the platform uses only its local records, while the personal agent receives a broader view of the same user.

We build these datasets from the 0-core population of Amazon Reviews 2023 \citep{hou2024blair}. We apply no additional $k$-core filter which preserves natural distributions of within-platform and cross-platform user histories.
Movie and Toy each contain 5,000 training, 1,000 validation, and 2,000 test episodes and provide seen-target-platform evaluation. Grocery and Beauty each contain 2,000 test episodes and are reserved for unseen-target-platform transfer.

\subsubsection{\textbf{OpenPlay: Real Cross-Platform Data}}

The Amazon construction provides scale and controlled transfer, but its platform boundary remains a proxy. For a real cross-platform setting, we use the OpenPlay dataset \citep{ballou2025openplay}, which links Steam, Nintendo, and Xbox activity through a shared pseudonymized person identifier. Steam is the target platform: earlier Steam titles form within-platform history, while earlier Nintendo titles and Xbox genres records from the same user form cross-platform history. The shared identifier gives direct person-level linkage across genuinely separate services without entity matching.

We require at least one within-platform and one cross-platform record but impose no additional activity threshold, similarly preserving the natural variation in gaming histories rather than restricting evaluation to highly active users. The resulting 645-user cohort is used only for external testing. It is excluded from agent training, hyperparameter selection, and checkpoint selection, providing an external test of whether mediation strategy learned on the Amazon proxy platforms transfers to a real cross-platform boundary.

\subsubsection{\textbf{Temporal Episodes and Candidate Sets}}

Each user contributes one temporally ordered episode. In Movie, Toy, Grocery, and Beauty, the user's final interaction on the target category is held out as $Y$ and treated as an implicit interaction target. In OpenPlay, $Y$ is the last-adopted Steam title that passes a positive-engagement threshold of 30 minutes of playtime, with adoption defined by first-play time. In all five datasets, each episode contains at least one within-platform record and one cross-platform record strictly before the target event.

The platform and personal agent operate over the same fixed set of 50 candidates: $Y$ and 49 items the user has not previously interacted with, sampled from the target platform. Negatives are sampled without replacement in proportion to item popularity, which favors frequently interacted items rather than rare long-tail items \citep{krichene2020sampled,ihemelandu2023candidate}. Candidates are randomly ordered for platform construction. After the platform-side recommender ranking, candidates are relabeled by platform rank, and this final representation is held fixed across all agent conditions. As all agents operate over the same candidate set, platform-relative changes reflect mediation of an available candidate set rather than different retrieval pools.
Table~\ref{tab:benchmark_stats} summarizes the resulting dataset sizes, history statistics, common cross-platform sources, and performance of the frozen target platform described next.

\begin{table}[t]
\centering
\caption{{MediateRec} statistics. Train/Val./Test report episode counts; Within/Cross are mean pre-target history lengths; Platform H@10 is the frozen platform's HR@10; Common cross-platform sources are ranked by user coverage.}
\label{tab:benchmark_stats}
\setlength{\tabcolsep}{3pt}
\small
\resizebox{\textwidth}{!}{
\begin{tabular}{lrrrrrrp{0.55\textwidth}}
\toprule
{MediateRec}
& Train
& Val.
& Test
& Within
& Cross
& Platform H@10
& Common Cross-Platform Sources
\\
\midrule

{Movie}
& 5,000
& 1,000
& 2,000
& 4.9
& 26.7
& 56.0
& Books (59.7\%); Home \& Kitchen (53.4\%); Electronics (50.3\%) \\

{Toy}
& 5,000
& 1,000
& 2,000
& 2.8
& 24.0
& 47.7
& Home \& Kitchen (71.3\%); Clothing, Shoes \& Jewelry (68.5\%); Electronics (52.4\%) \\

\midrule

{Grocery}
& --
& --
& 2,000
& 2.9
& 31.6
& 45.0
& Home \& Kitchen (77.7\%); Clothing, Shoes \& Jewelry (71.4\%); Health \& Household (60.3\%) \\

{Beauty}
& --
& --
& 2,000
& 3.2
& 26.6
& 48.6
& Clothing, Shoes \& Jewelry (74.1\%); Home \& Kitchen (71.4\%); Electronics (52.9\%) \\

{OpenPlay}
& --
& --
& 645
& 11.9
& 11.0
& 40.8
& Nintendo (85.1\%); Xbox (19.8\%) \\
\bottomrule
\end{tabular}
}
\end{table}


\subsection{Platform--Agent Interface}\label{sec:interface}

\subsubsection{\textbf{Target Platform Ranking}}

We construct a strong target-platform ranking so that generic repair of a weak candidate order is not mistaken for effective mediation. For each episode, a target-platform-trained SASRec model first ranks the 50 candidates using population interactions and the user's within-platform history \citep{kang2018sasrec}. This stage encodes population-level collaborative patterns unavailable to the personal agent. Following recent work on LLM-based recommendations and agentic recommendations \citep{hou2023llmrank,yue2023llamarec}, we use Claude Sonnet 4.6 to rerank the SASRec order at temperature 0, conditioned on the user's within-platform history. The reranking prompt is provided in Appendix~\ref{app:rerank}.
Cross-platform history is never exposed during target platform ranking construction, and the final returned platform ranking is frozen for all personal-agent training and evaluation.
The resulting platform rankings achieve mean HR@10 / NDCG@10 of $49.3\%/0.329$ across the four Amazon datasets and $40.8\%/0.241$ on OpenPlay, providing a strong default that the personal agent must preserve when correct and revise when cross-platform history provides stronger evidence.

\subsubsection{\textbf{Personal-Agent Mediation}}

The personal agent receives the platform proposal $B$, within-platform history $H^{\mathrm{within}}$, and cross-platform history $H^{\mathrm{cross}}$, but not platform scores, population interactions, or model states, matching the task's platform--agent information boundary. It then produces a ranked top-10 from the 50 candidates in a single call. We provide a default single-call mediation prompt in Appendix \ref{app:interface} that treats platform rank as meaningful evidence and asks the agent to revise it only when cross-platform history provides specific support. 
The interface therefore allows both deference to strong platform decisions and selective intervention when cross-platform history adds useful information.

\subsection{Evaluation Protocol}\label{sec:eval_protocol}

All selected users are disjoint across the target platforms and their train, validation, and test splits. Trained checkpoint selection uses only the seen-target-platform training and validation data. All held-out platforms are evaluated without adaptation.
All methods are evaluated on identical users, targets, candidates, and constructed platform rankings within each dataset. We report HR@\{3,5,10\} and NDCG@\{3,5,10\}, together with rescue, harmful override, and intervention, which is the number of platform-slate items replaced in the final slate, $d(S,P)=K-|S\cap P|$. 
The full-history condition provides $(H^{\mathrm{within}},H^{\mathrm{cross}})$, while the within-only condition uses $H^{-}=(H^{\mathrm{within}},\varnothing)$, holding the candidate set and platform proposal fixed. Movie and Toy are aggregated as seen target platforms, Grocery and Beauty as unseen target platforms, and OpenPlay is reported separately as the real cross-platform external test.

Appendix~\ref{app:benchmark} provides the further benchmark construction and implementation details.

\section{Personal Attribution Mediation Optimization}
\label{sec:pamo}

Outcome-only reinforcement learning can improve recommendation outcomes, but
it cannot distinguish a platform correction that genuinely depends on
cross-platform history from an equally successful correction produced by
generic reranking. Our {Personal Attribution Mediation Optimization
(PAMO)} addresses this ambiguity in three steps. First, it estimates how
strongly each sampled rationale depends on cross-platform history. Second, it
decomposes NDCG into rank-aware group-relative advantages. Third, it reallocates
the advantage assigned to platform-relative changes toward more strongly
supported responses while retaining their average platform-relative NDCG
magnitude. We then establish preservation and local-optimality properties of
this reallocation.

For each episode, PAMO samples $G$ responses from the frozen rollout policy:
\begin{equation}
(R_g,S_g)
\sim
\pi_{\bar\theta}(\cdot\mid B,H),
\qquad
g=1,\ldots,G,
\label{eq:pamo_rollouts}
\end{equation}
where $\pi_{\bar\theta}$ is fixed during the current update.

\subsection{Counterfactual Personal Mediation Support}
\label{sec:pamo_attribution}

Recommendation outcomes reveal whether a sampled slate succeeds, but not
whether the response actually relies on cross-platform history. To measure
this dependence, PAMO keeps the sampled response and platform proposal fixed
and removes only the cross-platform history. It then compares the likelihood
of the same response under the full-history and within-only inputs.

We score the visible rationale rather than the final identifier span. Once the
same rationale prefix is teacher-forced under both inputs, the final
identifiers are conditioned on that fixed prefix and can become less sensitive
to the removed history. The rationale more directly captures the
history-conditioned deliberation that produced the slate.

Let $R_g$ be the rationale generated before the \texttt{FINAL:}
marker, with $|R_g|$ scored tokens. We write the full-history input as
$(B,H)$ and the within-only input as $(B,H^{-})$, where
\[
H^{-}=(H^{\mathrm{within}},\varnothing)
\]
masks cross-platform history while retaining the complete within-platform
history. PAMO defines the \emph{personal mediation support} score
\begin{equation}
c_g
=
\frac{1}{|R_g|}
\log
\frac{
\pi_{\bar\theta}(R_g\mid B,H)
}{
\pi_{\bar\theta}(R_g\mid B,H^{-})
}.
\label{eq:personal_attribution}
\end{equation}

Because the full-history log-probabilities are retained during rollout
generation, computing $c_g$ requires one additional teacher-forced pass under
the within-only input. A larger $c_g$ means that the sampled rationale depends
more strongly on cross-platform history, conditional on the same platform
proposal and within-platform history. The score is model-relative: it measures
history dependence, not whether the history is correct or useful.
Recommendation outcomes provide that supervision. We normalize $c_g$ by
rationale length, stop gradients through it, and attach one score to the
response as a whole.

\subsection{Rank-Aware Platform-Relative Advantage}
\label{sec:pamo_rank}

For each ranking depth $k\in\{1,\ldots,K\}$, we call the top-$k$ boundary cutoff $k$, and a rank change affects exactly the cutoffs it crosses. Moving the target from rank 8 to rank 3, for example, changes the outcome at cutoffs $3$ through $7$. We use the exact decomposition of single-target NDCG@$K$ over these nested top-$k$ outcomes.

For rollout $g$ and the platform slate, define
\begin{equation}
z_{g,k}
=
\mathbf 1\{\operatorname{rank}_{S_g}(Y)\leq k\},
\qquad
z_{P,k}
=
\mathbf 1\{\operatorname{rank}_{P}(Y)\leq k\}.
\label{eq:cutoff_outcomes}
\end{equation}
Let $\ell_k$ denote the NDCG discount and $\alpha_k$ its marginal cutoff utility:
\begin{equation}
\ell_k
=
\frac{1}{\log_2(k+1)},
\qquad
\ell_{K+1}=0,
\qquad
\alpha_k
=
\ell_k-\ell_{k+1}.
\label{eq:ndcg_cutoff_weights}
\end{equation}
The rollout reward is exactly
\begin{equation}
r_g
\equiv
U_{\mathrm{NDCG}}(S_g,Y)
=
\sum_{k=1}^{K}
\alpha_k z_{g,k}.
\label{eq:ndcg_decomposition}
\end{equation}
Each $\alpha_k\geq0$ is the marginal NDCG gain from moving the target across cutoff $k$, so Eq.~\eqref{eq:ndcg_decomposition} lets PAMO model platform corrections at each affected cutoff without changing the underlying ranking utility.

At each cutoff, we use the group-relative centering of GRPO \citep{shao2024deepseekmath}. The rollout-group hit rate and centered advantage are
\begin{equation}
p_k
=
\frac{1}{G}
\sum_{g=1}^{G}
z_{g,k},
\qquad
A_{g,k}^{\mathrm{grp}}
=
z_{g,k}-p_k.
\label{eq:group_relative_advantage}
\end{equation}
There are $Gp_k$ successful rollouts, each with advantage $1-p_k$. Their total positive advantage, which also equals the magnitude of the total negative advantage, is
\begin{equation}
m_k
=
Gp_k(1-p_k).
\label{eq:cutoff_advantage_mass}
\end{equation}

Under uniform allocation, recombining the cutoff advantages recovers the centered NDCG advantage exactly, $\sum_{k}\alpha_k(z_{g,k}-p_k)=r_g-\bar r$ with $\bar r=\tfrac1G\sum_g r_g$. The decomposition therefore preserves the centered NDCG advantage of the matched GRPO baseline before support-based reallocation. Appendix~\ref{app:pamo_ndcg} gives the full decomposition and uniform-allocation
derivation.

PAMO modifies advantages only for responses whose top-$k$ outcome differs
from the platform. We define the \emph{platform-disagreement set}
\begin{equation}
\mathcal D_k
=
\{g:z_{g,k}\neq z_{P,k}\},
\qquad
s_k
=
1-2z_{P,k}.
\label{eq:platform_disagreement}
\end{equation}
If the platform misses at cutoff $k$ ($s_k=1$), $\mathcal D_k$ contains
top-$k$ rescues. If the platform succeeds ($s_k=-1$), it contains harmful
overrides that move the target below the cutoff. Preserved hits and mutual misses stay outside $\mathcal D_k$ and keep the standard group-relative advantage, as they are not platform-relative interventions at that cutoff. We call a cutoff \emph{active} when $0<p_k<1$. At $p_k\in\{0,1\}$, Eq.~\eqref{eq:cutoff_advantage_mass} gives $m_k=0$ and the cutoff contributes no update, so PAMO solves the allocation only at active cutoffs.

\subsection{Value-Preserving Advantage Reallocation}
\label{sec:pamo_reweighting}

\subsubsection{\textbf{Value-Constrained Allocation}}

Within each active platform disagreement set, PAMO tilts the advantage allocation toward more personally supported responses while not lowering its average platform-relative recommendation value. For $g\in\mathcal D_k$, define the \emph{direction-aligned NDCG magnitude} as
\begin{equation}
r_P
=
U_{\mathrm{NDCG}}(P,Y),
\qquad
v_{g,k}
=
s_k(r_g-r_P).
\label{eq:mediation_value_score}
\end{equation}
For every $g\in\mathcal D_k$, the sign $s_k$ makes
$v_{g,k}=|r_g-r_P|\geq0$. Thus, $v_{g,k}$ is larger for a stronger rescue
when the platform misses and for a more severe harmful override when the
platform succeeds. The cutoff determines whether the response is a rescue
or harmful override, while $v_{g,k}$ measures the NDCG magnitude of the
full rank change.

Let $n_k=|\mathcal D_k|$ and let $u_k$ be the uniform distribution over this set:
\begin{equation}
u_{g,k}
=
\frac{1}{n_k},
\qquad
\bar v_k
=
\sum_{g\in\mathcal D_k}
u_{g,k}v_{g,k}.
\label{eq:uniform_mediation_value}
\end{equation}
Reallocating advantage by personal support alone can concentrate it on a highly history-dependent but low-value rescue, or place too little negative advantage on a severe harmful override. PAMO therefore requires the reallocated advantage to retain at least the average direction-aligned NDCG magnitude of the original uniform allocation, a constraint we call the \emph{value floor}. 

Using the value floor rather than a fixed weighted sum permits support-based reallocation only when it does not reduce the original average value. Here $\eta\geq0$ is the support strength. For each active $\mathcal D_k$, PAMO chooses
\begin{equation}
\begin{aligned}
w_k^\star
=
\arg\max_{w_k\in\mathbb R_{+}^{n_k},\;\sum_{g}w_{g,k}=1}
\quad&
\eta
\sum_{g\in\mathcal D_k}
w_{g,k}c_g
-
\operatorname{KL}(w_k\Vert u_k)
\\
\text{subject to}\quad&
\sum_{g\in\mathcal D_k}
w_{g,k}v_{g,k}
\geq
\bar v_k.
\end{aligned}
\label{eq:pamo_allocation}
\end{equation}
The objective tilts the advantage allocation toward personal support while limiting departure from the uniform group-relative allocation, and the value floor excludes allocations with lower average direction-aligned NDCG magnitude. Because $u_k$ is feasible (it meets the value floor with equality) and the negative-KL objective is strictly concave on the simplex, the solution is unique.

The unique optimizer has the dual-parameterized exponential form
\begin{equation}
w_{g,k}^\star
=
\frac{
\exp\!\left(
\eta c_g+\lambda_k v_{g,k}
\right)
}{
\sum_{j\in\mathcal D_k}
\exp\!\left(
\eta c_j+\lambda_k v_{j,k}
\right)
},
\qquad
g\in\mathcal D_k,
\label{eq:pamo_weights}
\end{equation}
where $\lambda_k\geq0$ is the dual coefficient of the value floor.
When the unconstrained support tilt already satisfies the floor,
$\lambda_k=0$. Otherwise, $\lambda_k$ is the unique value that makes the
constraint in Eq.~\eqref{eq:pamo_allocation} active.
Appendix~\ref{app:pamo_dual} derives Eq.~\eqref{eq:pamo_weights} and the one-dimensional $\lambda_k$ search.

\subsubsection{\textbf{PAMO Advantage}}
PAMO assigns the cutoff-level advantage
\begin{equation}
A_{g,k}^{\mathrm{PAMO}}
=
\begin{cases}
s_k m_k w_{g,k}^\star,
&
g\in\mathcal D_k,
\\[3pt]
A_{g,k}^{\mathrm{grp}},
&
g\notin\mathcal D_k.
\end{cases}
\label{eq:pamo_cutoff_advantage}
\end{equation}
The allocation changes only which platform-disagreement rollouts receive the fixed signed advantage mass $s_k m_k$. Rollouts whose cutoff outcome agrees with the platform retain the group-relative advantage in Eq.~\eqref{eq:group_relative_advantage}.

Because the scale of $c_g$ changes during training, we calibrate the support
strength using a dimensionless concentration parameter $\kappa$. Let
$\widehat{\sigma}_c$ be the mean, over all active platform-disagreement sets
in the current update, of the within-set standard deviation of $c_g$. We set
\begin{equation}
\eta
=
\min\left\{
\frac{\kappa}{\max(\widehat{\sigma}_c,\epsilon_c)},
\eta_{\max}
\right\},
\label{eq:pamo_kappa}
\end{equation}
where $\epsilon_c>0$ is a numerical floor and $\eta_{\max}$ caps the support
strength. Thus, $\kappa$ controls the typical spread of the support logits:
$\kappa=0$ gives the uniform allocation, while larger values make the
allocation more responsive to differences in personal mediation support.

Finally, the cutoff advantages are combined into one response advantage using the same marginal utilities that define NDCG,
\begin{equation}
A_g^{\mathrm{PAMO}}
=
\sum_{k=1}^{K}
\alpha_k
A_{g,k}^{\mathrm{PAMO}},
\label{eq:pamo_rollout_advantage}
\end{equation}
which is used in the same clipped token-level GRPO objective as the matched outcome-only baseline \citep{shao2024deepseekmath}. PAMO adds one masked teacher-forced scoring pass during training, with no additional rollouts and no inference-time component. Appendix~\ref{app:pamo_training} gives the clipped token-level objective and computational details.

\subsection{Theoretical Analysis}
\label{sec:pamo_theory}

The allocation in Eq.~\eqref{eq:pamo_allocation} reallocates advantage within each platform disagreement set while preserving its cutoff-level mass.

\begin{proposition}[Advantage-Mass and Value Preservation]
\label{prop:pamo_preservation}
Fix an active cutoff ($0<p_k<1$) and support strength $\eta\geq0$, and let $w_k^\star$ solve Eq.~\eqref{eq:pamo_allocation}. PAMO preserves the aggregate positive and negative group-relative advantage mass:
\begin{equation}
\sum_{g:z_{g,k}=1}
A_{g,k}^{\mathrm{PAMO}}
=
m_k,
\qquad
\sum_{g:z_{g,k}=0}
A_{g,k}^{\mathrm{PAMO}}
=
-m_k.
\label{eq:advantage_preservation}
\end{equation}
Its allocation also satisfies
\begin{equation}
\mathbb E_{w_k^\star}[v_k]
\geq
\mathbb E_{u_k}[v_k],
\qquad
\mathbb E_{w_k^\star}[c]
\geq
\mathbb E_{u_k}[c].
\label{eq:value_attribution_guarantees}
\end{equation}
\end{proposition}

\paragraph{{Proof sketch.}} At an active cutoff, $\mathcal D_k$ contains exactly one side of the binary group outcome---rescues when the platform misses and harmful overrides when it succeeds. Since $w_k^\star$ sums to one, PAMO redistributes but does not change the signed advantage mass $s_k m_k$, and the platform-agreement side is untouched. The value inequality is the feasibility constraint, and comparing the optimum with the feasible uniform allocation $u_k$ gives $\mathbb E_{w_k^\star}[c]\geq\mathbb E_{u_k}[c]$. Appendix~\ref{app:proof_preservation} gives the full proof.

To characterize the local reallocation direction, center the support score and value magnitude under the uniform distribution:
\begin{equation}
\widetilde c_{g,k}
=
c_g-\mathbb E_{u_k}[c],
\qquad
\widetilde v_{g,k}
=
v_{g,k}-\mathbb E_{u_k}[v_k].
\label{eq:centered_attribution_value}
\end{equation}
Define the cone of admissible first-order weight reallocations
\begin{equation}
\mathcal K_k
=
\left\{
q\in\mathbb R^{n_k}:
\mathbf 1^\top q=0,\;
\langle q,\widetilde v_k\rangle\geq0
\right\},
\qquad
q_k
=
\operatorname{Proj}_{\mathcal K_k}
(\widetilde c_k).
\label{eq:admissible_projection}
\end{equation}

\begin{theorem}[Local Optimality of the PAMO Allocation Direction]
\label{thm:pamo_local}
Fix an active cutoff $k$ with $n_k\geq2$. The right derivative of the PAMO weights at the uniform allocation is
\begin{equation}
\left.
\frac{d w_k^\star}{d\eta}
\right|_{\eta=0^{+}}
=
\frac{1}{n_k}q_k.
\label{eq:pamo_weight_first_order}
\end{equation}
When $q_k\neq0$, the direction $q_k/\|q_k\|_2$ maximizes the first-order increase in average personal mediation support among unit-Euclidean-norm infinitesimal reallocations that preserve total weight and do not decrease average direction-aligned NDCG magnitude.
\end{theorem}

\paragraph{{Proof sketch.}} Expanding the exponential-family solution around the uniform allocation, $\lambda_k(\eta)=\eta\mu_k+o(\eta)$. The simplex constraint centers the support score, and an active value floor removes the component of the centered support direction that would decrease the value magnitude. Because the Hessian of $\operatorname{KL}(w_k\Vert u_k)$ is isotropic on the simplex tangent space at $u_k$, the resulting direction is the Euclidean projection of $\widetilde c_k$ onto $\mathcal K_k$, whose optimality condition gives the maximal first-order support gain among unit-norm admissible reallocations. Appendix~\ref{app:proof_local} gives the full proof, including existence of
the one-sided derivative.

If the centered support direction is already value-aligned then $q_k=\widetilde c_k$. Otherwise the projection in Eq.~\eqref{eq:admissible_projection} removes only the value-conflicting component. Thus, at each active cutoff, PAMO follows the steepest support-increasing first-order advantage reallocation that preserves total advantage mass and does not decrease the average direction-aligned NDCG magnitude. These results characterize advantage allocation within a fixed sampled rollout group, treating $c_g$ and $v_{g,k}$ as fixed, and do not imply global optimality of the neural policy or a guaranteed held-out utility improvement.

\section{Experiments}
\label{sec:experiment}

The experiments are designed to answer the following research questions on personal-agent mediated recommendation:

\begin{itemize}
    \item[\textbf{RQ1.}] Can personal-agent mediation improve a strong platform proposal, and does PAMO outperform matched RL training baselines?
    \item[\textbf{RQ2.}] How much of the mediation gain comes from cross-platform user history rather than generic reranking?
    \item[\textbf{RQ3.}] How do the value floor and different values of support concentration $\kappa$ affect PAMO?
    \item[\textbf{RQ4.}] How does PAMO use cross-platform history to decide when to revise or preserve the platform ranking?
\end{itemize}

\subsection{Experimental Setup}
\label{sec:experimental_setup}

\subsubsection{\textbf{Models and Baselines}}

We report the frozen Platform ranking as the no-mediation baseline. As described in Section~\ref{sec:interface}, it is produced by a target-platform SASRec model followed by Claude Sonnet 4.6 reranking using within-platform history only \citep{kang2018sasrec,hou2023llmrank,yue2023llamarec}. This baseline combines population-level collaborative signals with LLM reranking and defines the platform proposal received by every personal agent. Rescue, harmful override, and intervention are measured relative to it. Existing platform-side agentic recommenders operate upstream by constructing or refining such a proposal, rather than at the personal-agent decision point studied here.

All trainable personal agents use Qwen3-4B-Instruct-2507 as the backbone. Claude Sonnet 4.6 supplies the supervised demonstrations for behavior initialization. We additionally evaluate Claude Haiku 4.5, Sonnet 4.6, and Opus 4.6 as proprietary inference-time reference models. All personal-agent methods receive the same frozen platform proposal and use the same information interface.

We compare the base model, supervised fine-tuning (SFT), GRPO \citep{shao2024deepseekmath}, and PAMO. GRPO is the matched outcome-only RL baseline and uses NDCG@10 as its reward. It shares PAMO's SFT initialization, training data, rollout budget, optimizer, and update budget, differing only in advantage reallocation. For ablation, we also evaluate \emph{PAMO w/o Value Floor}, which removes the value floor in Section~\ref{sec:pamo_reweighting} and reallocates advantage using personal mediation support alone.

\subsubsection{\textbf{Training}}

We initialize the personal agent on 2,000 Sonnet demonstrations, 1,000 each from Movie and Toy, retained without filtering by recommendation outcome for format learning. SFT uses three epochs, batch size 64, maximum sequence length 8,192, and learning rate $10^{-5}$. All RL methods start from the same SFT checkpoint. Each update uses 64 prompts and $G=8$ rollouts per prompt, sampled with temperature 1.0 and top-$p$ 1.0, with learning rate $10^{-6}$, PPO minibatch size 32, clipping threshold $\epsilon=0.2$, and no KL penalty. Prompt and response lengths are capped at 8,192 and 2,048 tokens, and group advantages are not standardized. Main runs use 400 updates with validation and checkpointing every 50 updates. 
PAMO computes personal mediation support on the visible rationale and masks cross-platform history, sweeping $\kappa\in\{0.25,0.5,0.75\}$ on the validation set with the effective inverse temperature capped at $\eta_{\max}=60$. We implement training with VeRL \citep{sheng2024hybridflow} and use vLLM \citep{kwon2023efficient} for rollout generation.

\subsubsection{\textbf{Evaluation}} We select the checkpoint with the highest validation NDCG@10. At test time, each model produces one response per episode using temperature $0.7$, top-$p=0.8$, and top-$k=20$. We extract the candidate identifiers following the \texttt{FINAL:} marker, remove duplicates, and use their best-first order as the final slate. We report HR@\{3,5,10\} and NDCG@\{3,5,10\}. At $K=10$, \emph{rescue} is the percentage of episodes in which the platform misses the target but the personal agent recovers it, while \emph{harmful override} is the percentage in which the platform includes the target but the agent removes it. Mean intervention is the number of platform top-10 items replaced in the final slate. 

\subsection{Overall Mediation Performance (RQ1)}

\subsubsection{\textbf{Access to Cross-Platform History Is Insufficient}}

Tables~\ref{tab:domain_results} and~\ref{tab:openplay_results} report mediation results across the five {MediateRec} datasets. Access to cross-platform history alone does not produce effective mediation. The base Qwen3-4B-Instruct falls below the platform on all five datasets,
and its harmful-override rate exceeds its rescue rate throughout. 
Cross-platform history therefore needs to be combined with selective deference to the platform proposal. Among the proprietary models, Haiku only nearly match the platform even with cross-platform history, whereas Sonnet and Opus improve HR@10 over the platform on all five datasets. Their remaining harmful-override rates show that stronger model capability alone does not solve selective mediation.

\begin{table}[t]
\centering
\caption{Recommendation and mediation results on the four proxy cross-platform test sets.
H@X and N@X denote HR@X and NDCG@X. Best results are in \textbf{bold};
second-best results are \underline{underlined}.}
\vspace{-0.25em}
\label{tab:domain_results}
\setlength{\tabcolsep}{3pt}
\resizebox{\textwidth}{!}{
\begin{tabular}{l*{9}{c}@{\hspace{6pt}}*{9}{c}}
\toprule
& \multicolumn{9}{c}{{MediateRec-Movie}}
& \multicolumn{9}{c}{{MediateRec-Toy}} \\
\cmidrule(lr){2-10}
\cmidrule(lr){11-19}
Method
& H@3 & H@5 & H@10 & N@3 & N@5 & N@10 & Resc. & Harm & Int.
& H@3 & H@5 & H@10 & N@3 & N@5 & N@10 & Resc. & Harm & Int. \\
\midrule
Platform
& 37.0 & 44.5 & 56.0 & 0.319 & 0.349 & 0.386 & -- & -- & --
& 31.2 & 36.8 & 47.7 & 0.268 & 0.291 & 0.326 & -- & -- & -- \\

Claude Haiku 4.5
& 35.3 & 42.4 & 56.8 & 0.304 & 0.333 & 0.380 & 10.9 & 10.1 & 3.77
& 30.4 & 37.3 & 48.2 & 0.261 & 0.290 & 0.325 & 9.6 & 9.1 & 3.96 \\

Claude Sonnet 4.6
& 40.4 & \underline{49.0} & \underline{63.1} & 0.340 & 0.376 & 0.421 & \textbf{12.2} & 5.0 & 3.27
& 32.6 & 39.5 & 52.3 & 0.281 & 0.308 & 0.350 & \textbf{12.1} & 7.5 & 3.71 \\

Claude Opus 4.6
& \textbf{42.1} & \textbf{50.2} & \textbf{63.2} & \textbf{0.353} & \textbf{0.387} & \textbf{0.428} & \underline{11.6} & 4.2 & 2.92
& 33.5 & 40.4 & 52.4 & 0.285 & 0.313 & 0.352 & 11.7 & 7.0 & 3.62 \\

\midrule
Qwen3-4B-Instruct
& 34.1 & 41.0 & 50.1 & 0.298 & 0.326 & 0.355 & 4.1 & 9.9 & 2.90
& 29.4 & 34.8 & 45.5 & 0.254 & 0.276 & 0.311 & 5.5 & 7.7 & 3.14 \\

$\;\;+$SFT
& 37.5 & 45.0 & 57.7 & 0.318 & 0.349 & 0.390 & 8.2 & 6.5 & 2.93
& 30.6 & 37.4 & 48.6 & 0.265 & 0.293 & 0.329 & 8.7 & 7.8 & 3.43 \\

$\;\;+$GRPO
& 38.6 & 47.4 & 61.2 & 0.328 & 0.364 & 0.408 & 9.2 & \textbf{3.9} & 2.71
& \underline{33.9} & \underline{41.0} & \underline{53.4} & \underline{0.286} & \underline{0.315} & \underline{0.355} & 11.2 & \underline{5.6} & 3.64 \\

$\;\;+$\textbf{PAMO}
& \underline{41.0} & \underline{49.0} & 62.2 & \underline{0.351} & \underline{0.384} & \underline{0.426} & 10.3 & \underline{4.0} & 2.76
& \textbf{35.0} & \textbf{42.2} & \textbf{54.8} & \textbf{0.302} & \textbf{0.332} & \textbf{0.372} & \underline{11.9} & \textbf{4.8} & 3.31 \\

\midrule
& \multicolumn{9}{c}{{MediateRec-Grocery}}
& \multicolumn{9}{c}{{MediateRec-Beauty}} \\
\cmidrule(lr){2-10}
\cmidrule(lr){11-19}
Method
& H@3 & H@5 & H@10 & N@3 & N@5 & N@10 & Resc. & Harm & Int.
& H@3 & H@5 & H@10 & N@3 & N@5 & N@10 & Resc. & Harm & Int. \\
\midrule
Platform
& 27.6 & 33.6 & 45.0 & 0.238 & 0.263 & 0.299 & -- & -- & --
& 27.4 & 34.4 & 48.6 & 0.230 & 0.259 & 0.305 & -- & -- & -- \\

Claude Haiku 4.5
& 26.4 & 32.9 & 43.0 & 0.224 & 0.251 & 0.283 & 8.9 & 11.1 & 4.08
& 27.4 & 34.5 & 47.3 & 0.227 & 0.256 & 0.297 & 8.4 & 9.8 & 3.73 \\

Claude Sonnet 4.6
& 28.3 & \underline{36.0} & \underline{49.1} & 0.239 & \underline{0.271} & 0.313 & \textbf{11.0} & 7.0 & 3.84
& 30.2 & 37.9 & 50.6 & \underline{0.248} & \underline{0.279} & 0.320 & \underline{10.4} & 8.4 & 3.81 \\

Claude Opus 4.6
& 27.3 & 35.6 & 48.5 & 0.232 & 0.266 & 0.308 & 9.8 & 6.3 & 3.29
& \underline{30.8} & \underline{38.1} & \textbf{51.9} & \underline{0.248} & 0.278 & \underline{0.322} & \textbf{11.6} & 8.3 & 3.75 \\

\midrule
Qwen3-4B-Instruct
& 26.0 & 30.8 & 41.0 & 0.225 & 0.244 & 0.277 & 4.7 & 8.8 & 3.09
& 25.8 & 31.9 & 41.5 & 0.217 & 0.243 & 0.273 & 4.7 & 11.8 & 3.67 \\

$\;\;+$SFT
& 27.2 & 33.6 & 43.4 & 0.230 & 0.256 & 0.288 & 6.8 & 8.4 & 3.35
& 27.8 & 33.9 & 46.8 & 0.229 & 0.254 & 0.296 & 6.6 & 8.4 & 2.79 \\

$\;\;+$GRPO
& \underline{28.8} & 35.4 & 48.8 & \underline{0.244} & \underline{0.271} & \underline{0.314} & 8.4 & \underline{4.7} & 2.87
& 29.4 & 37.8 & 51.2 & 0.241 & 0.276 & 0.319 & 8.9 & \underline{6.3} & 3.19 \\

$\;\;+$\textbf{PAMO}
& \textbf{30.4} & \textbf{37.5} & \textbf{50.6} & \textbf{0.264} & \textbf{0.292} & \textbf{0.335} & \underline{10.0} & \textbf{4.5} & 2.96
& \textbf{31.1} & \textbf{39.1} & \underline{51.7} & \textbf{0.260} & \textbf{0.293} & \textbf{0.333} & 8.6 & \textbf{5.5} & 3.05 \\

\bottomrule
\end{tabular}
}
\end{table}

\subsubsection{\textbf{PAMO Improves Outcome-Only Training}}

Among the training methods, SFT followed by GRPO progressively recovers selective mediation, and PAMO improves further over the matched GRPO baseline. PAMO improves both HR@10 and NDCG@10 over GRPO on all five datasets. 
The gain does not come from more aggressive intervention: mean intervention stays similar, 
\begin{wraptable}{r}{0.515\columnwidth}
\centering
\caption{Recommendation and mediation results on the real cross-platform test set.}
\label{tab:openplay_results}
\setlength{\tabcolsep}{2.2pt}
\resizebox{\linewidth}{!}{
\begin{tabular}{lccccccccc}
\toprule
& \multicolumn{9}{c}{{MediateRec-OpenPlay}} \\
\cmidrule(lr){2-10}
Method
& H@3 & H@5 & H@10
& N@3 & N@5 & N@10
& Resc. & Harm & Int. \\
\midrule

Platform
& 20.5 & 27.3 & 40.8
& 0.169 & 0.198 & 0.241
& -- & -- & -- \\

Claude Haiku 4.5
& 20.5 & 28.2 & 40.8
& 0.160 & 0.191 & 0.231
& 12.2 & 12.2 & 4.52 \\

Claude Sonnet 4.6
& \underline{23.7} & 30.5 & 44.0
& \underline{0.180} & 0.208 & 0.251
& 13.5 & 10.2 & 4.40 \\

Claude Opus 4.6
& 21.1 & 29.9 & 45.4
& 0.174 & 0.211 & 0.260
& \underline{13.6} & 9.0 & 3.75 \\

Qwen3-4B-Instruct
& 20.2 & 28.2 & 38.9
& 0.165 & 0.198 & 0.232
& 8.5 & 10.4 & 3.58 \\

$\;\;+$SFT
& 19.5 & 26.0 & 39.7
& 0.160 & 0.187 & 0.231
& 10.5 & 11.6 & 3.97 \\

$\;\;+$GRPO
& 22.2 & \underline{30.7} & \underline{47.4}
& 0.179 & \underline{0.214} & \underline{0.267}
& \textbf{13.8} & \underline{7.1} & 3.40 \\

$\;\;+$\textbf{PAMO}
& \textbf{26.2} & \textbf{32.7} & \textbf{47.9}
& \textbf{0.208} & \textbf{0.235} & \textbf{0.284}
& 13.2 & \textbf{6.0} & 3.40 \\

\bottomrule
\end{tabular}
}
\end{wraptable}
while PAMO reduces harmful overrides on four datasets and improves the rescue--harm balance overall. Using a 4B open model, PAMO is competitive with the proprietary models and exceeds them on many unseen-target and OpenPlay metrics. On {MediateRec-OpenPlay}, evaluated without any Open Play training or model selection, PAMO obtains the highest ranking metrics. Relative to GRPO, its largest gains occur near the top of the ranking, with higher H@3 and NDCG@10 and a lower harmful-override rate.

\subsection{Cross-Platform History Contribution (RQ2)} 
\begin{figure*}[t]
\centering
\includegraphics[width=\textwidth]{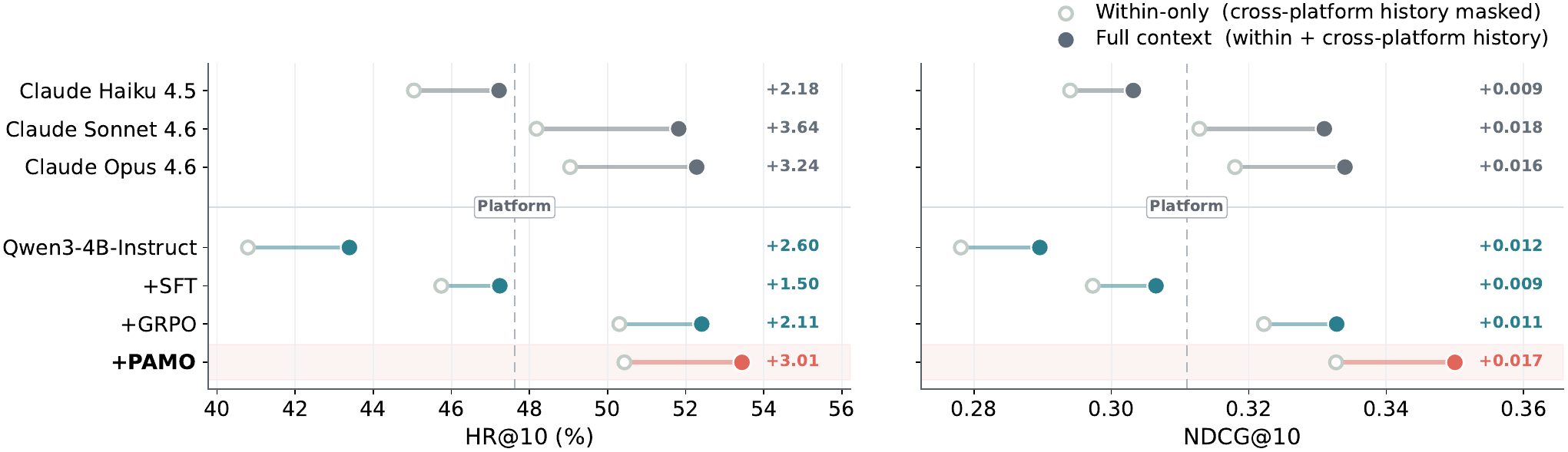}
\caption{Effect of cross-platform history averaged over all five test sets in {MediateRec}. Open markers use within-platform history only and filled markers use the full-context input. The dashed vertical line marks the frozen platform baseline.}
\label{fig:crosshist}
\end{figure*}

To isolate the contribution of cross-platform history, we re-evaluate each model after masking only the cross-platform records. The candidate set, platform proposal, and within-platform history remain unchanged, so the comparison removes the information unavailable to the platform without changing the underlying recommendation episode. Figure~\ref{fig:crosshist} reports the five-dataset macro average. Adding cross-platform history improves both HR@10 and NDCG@10 for every evaluated model, showing that the mediation gains cannot be explained by generic reranking alone. The largest improvements occur for Sonnet, Opus, and PAMO, indicating that stronger models and targeted post-training make better use of the same cross-platform evidence. Together with the base-model results in RQ1, this finding also shows that access to cross-platform history is useful but insufficient by itself: the agent must learn when that evidence is specific enough to justify changing the platform ranking.

\subsection{Ablation and Sensitivity Analysis (RQ3)}

\subsubsection{\textbf{Effect of Support Concentration $\kappa$}}

\begin{wrapfigure}{r}{0.5\columnwidth}
\centering
\includegraphics[width=\linewidth]{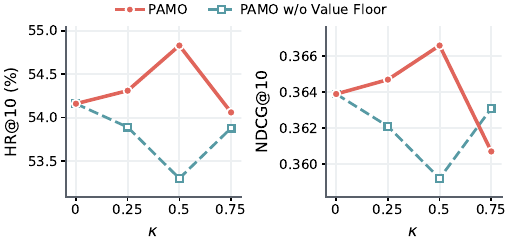}
\caption{Effect of support concentration $\kappa$ and the value floor,
averaged over the four proxy cross-platform test sets.}
\vspace{-2em}
\label{fig:kappa}
\end{wrapfigure}

The support concentration parameter $\kappa$ controls how strongly PAMO reallocates advantage toward responses with higher personal mediation support. Figure~\ref{fig:kappa} shows an inverted-U pattern. At $\kappa=0$, personal mediation support does not affect the allocation. 
Increasing $\kappa$ to $0.5$ improves both HR@10 and NDCG@10, 
showing that the counterfactual support signal provides useful information beyond the ranking outcome. Performance weakens at the largest tested value, $\kappa=0.75$, indicating that allowing support differences to dominate the allocation is less effective. We therefore use $\kappa=0.5$ in the main experiments.

\subsubsection{\textbf{Role of the Value Floor}} 

We next remove the value floor while retaining the same personal mediation support signal. This variant reallocates advantage according to history dependence alone, without requiring the allocation to preserve average platform-relative NDCG magnitude. As shown in Figure~\ref{fig:kappa}, it performs below the $\kappa=0$ baseline across the tested support strengths, whereas full PAMO improves both HR@10 and NDCG@10. This contrast shows why personal mediation support cannot serve as a complete training signal by itself. A response can depend strongly on cross-platform history while still producing a weak rescue or a severe harmful override. The value floor prevents such responses from receiving a favorable reallocation merely because they are history-dependent, keeping the support signal aligned with recommendation quality.

\subsection{PAMO Mediation Behavior Analysis (RQ4)}

\subsubsection{\textbf{Aggregate Mediation Behavior}}

We first examine how PAMO uses cross-platform history when it successfully
overrides the platform. We annotate the visible rationales of the 900 test
episodes in which the platform top-10 misses the target and PAMO recovers it.
Using Claude Sonnet 4.6 as the annotator over each rationale and its corresponding
cross-platform records, we identify the primary evidence behind each rescue.

\begin{wrapfigure}{r}{0.5\columnwidth}
\centering
\includegraphics[width=\linewidth]{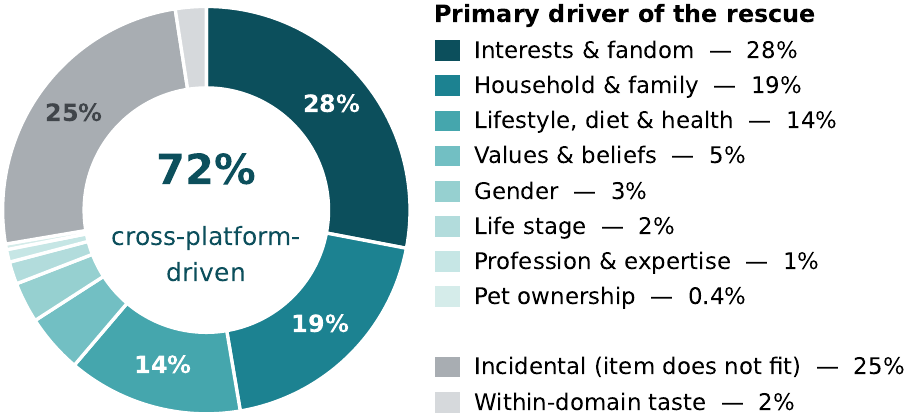}
\caption{Primary evidence behind PAMO's 900 rescues.}
\label{fig:persona}
\end{wrapfigure}

Figure~\ref{fig:persona} shows that cross-platform user context is the primary
driver in $72\%$ of rescues, rather than incidental reranking or
within-platform taste alone. The most common forms are interests and fandom
($28\%$), household and family ($19\%$), and lifestyle, diet, and health
($14\%$). The remaining rescues are mainly incidental ($25\%$), with only a
small fraction attributed primarily to within-platform taste. Thus, most PAMO
rescues are explicitly motivated by user context derived from records
unavailable to the target platform.

\definecolor{caseteal}{HTML}{2A7F8E}
\definecolor{caseamber}{HTML}{B5722A}

\newcolumntype{Y}{>{\raggedright\arraybackslash}X}

\newtcolorbox{casebox}[2]{
  enhanced,
  arc=2pt,
  boxrule=0.6pt,
  colframe=#2,
  colback=#2!3,
  colbacktitle=#2,
  coltitle=white,
  fonttitle=\bfseries\footnotesize,
  title={#1},
  fontupper=\footnotesize,
  fontlower=\footnotesize,
  colbacklower=#2!10,
  left=4pt,
  right=4pt,
  top=3pt,
  bottom=3pt,
}

\begin{figure}[t]
\centering

\begin{tcbraster}[
  raster columns=2,
  raster equal height=rows,
  raster column skip=5pt
]

\begin{casebox}
{Case 1\, $\cdot$\, Intervene -- MediateRec-Movie}
{caseteal}

{\setlength{\tabcolsep}{0pt}
\renewcommand{\arraystretch}{1.18}
\begin{tabularx}{\linewidth}{
  @{}
  >{\bfseries\raggedright\arraybackslash}p{3.8em}
  @{\hspace{2pt}}
  Y
  @{}
}

Target
&
\emph{Robert Plant and the Strange Sensation}
\\
&
\mbox{\colorbox{caseteal!15}
{\scriptsize\textbf{rank $\#37\!\rightarrow\!\#1$}}}
\\[1pt]

Within
&
\emph{Dead Poets Society}\,$\star$5,\,
\emph{American Sniper}\,$\star$5
\\
&
\textcolor{gray}{\scriptsize no music signal}
\\[1pt]

Cross
&
\emph{Led Zeppelin IV}\,$\star$4,\,
\emph{The Complete BBC Sessions}\,$\star$5
\\

\end{tabularx}}

\tcblower

\textbf{PAMO rationale.}
``\dots{} C37 (Robert Plant): Led Zeppelin IV + BBC Sessions
= strong specific override.''

\textbf{Decision.}
specific cross-platform evidence
$\Rightarrow$ \textbf{override the platform}.

\end{casebox}
\begin{casebox}
{Case 2\, $\cdot$\, Defer -- MediateRec-Toy}
{caseamber}

{\setlength{\tabcolsep}{0pt}
\renewcommand{\arraystretch}{1.18}
\begin{tabularx}{\linewidth}{
  @{}
  >{\bfseries\raggedright\arraybackslash}p{3.8em}
  @{\hspace{2pt}}
  Y
  @{}
}

Target
&
\emph{Zelda: Twilight Princess Link} figure
\\
&
\mbox{\colorbox{caseamber!15}
{\scriptsize\textbf{rank $\#1$ kept}}}
\\[1pt]

Within
&
\emph{Code Geass Lancelot} figure\,$\star$3
\\[1pt]

Cross
&
\emph{Magic Bullet blender}\,$\star$2,\,
putting mat\,$\star$1,\,
\emph{Overwatch} comic\,$\star$4
\\

\end{tabularx}}

\tcblower

\textbf{PAMO rationale.}
``Magic Bullet Blender---home appliance, not relevant to toy selection
\dots{} Putting mat---golf interest (weak signal).''

\textbf{Decision.}
no specific cross-platform support
$\Rightarrow$ \textbf{keep the platform ranking}.

\end{casebox}

\end{tcbraster}

\caption{Representative PAMO rationales illustrating intervention under
specific cross-platform evidence and deference under weak cross-platform evidence.}
\label{fig:cases}

\end{figure}

\subsubsection{\textbf{Intervening versus Deferring}}

Aggregate categories do not show how PAMO turns cross-platform history into a mediation decision. Figure~\ref{fig:cases} therefore presents two representative visible rationales illustrating the two behaviors central to the task: intervening when cross-platform evidence is specific and deferring when it is weak.
In the first case, 
the within-platform history provides no music signal, while the cross-platform history contains \emph{Led Zeppelin IV} and \emph{The Complete BBC Sessions}. PAMO explicitly connects these records to Robert Plant and promotes his concert film from platform rank 37 to rank 1. In the second case, the platform ranks a \emph{Zelda} collectible first, while the cross-platform history provides only weak or unrelated evidence. PAMO recognizes this lack of support and preserves the platform's top item. Together, the cases illustrate the selective behavior targeted by PAMO: cross-platform history can justify a substantial intervention when its support is specific, while weak evidence leads to deference.

\section{Conclusion}

In this paper, we formalize the {Personal-Agent Mediated Recommendation} task, in which a user-authorized personal agent uses cross-platform history to mediate a platform proposal and produce the final slate. We introduce {MediateRec}, which combines scalable proxy cross-platform environments with a real cross-platform external test. We also propose PAMO, which estimates personal mediation support through counterfactual masking and performs value-preserving advantage reallocation. Across seen and unseen proxy target platforms and the real cross-platform test, PAMO improves over matched outcome-only RL while achieving a stronger rescue--harm balance. These results show that effective mediation requires not only access to cross-platform user history, but also learning when that history should override population evidence encoded by a strong platform ranking.

\clearpage
\newpage
\bibliographystyle{assets/plainnat}
\bibliography{main}

@article{lin2026usergoverned,
  title={LLM Agents Enable User-Governed Personalization Beyond Platform Boundaries},
  author={Lin, Jiacheng and Qian, Kun and Srinivasan, Arvind and Wang, Tian and Han, Fang and Hu, Changran and Liu, Junze and Wang, Ziyi and Xu, Hanwen and Xue, Mengmeng and others},
  journal={arXiv preprint arXiv:2605.09794},
  year={2026}
}

@misc{meta2026muse,
  author = {{Meta}},
  title  = {Introducing Muse: The World's First Personal AI Agent Built for Everyone},
  year   = {2026},
  month  = sep,
  url    = {https://about.fb.com/news/2026/09/introducing-muse-personal-ai-agent/}
}

@article{liu2026governable,
  title={From hidden profiles to governable personalization: Recommender systems in the age of llm agents},
  author={Liu, Jiahao and Han, Mingzhe and Liu, Guanming and Wang, Weihang and Li, Dongsheng and Gu, Hansu and Zhang, Peng and Lu, Tun and Gu, Ning},
  journal={arXiv preprint arXiv:2604.20065},
  year={2026}
}

@inproceedings{xia2026macf,
author = {Xia, Yu and Kim, Sungchul and Yu, Tong and Rossi, Ryan A. and McAuley, Julian},
title = {Multi-Agent Collaborative Filtering: Orchestrating Users and Items for Agentic Recommendations},
year = {2026},
isbn = {9798400723070},
publisher = {Association for Computing Machinery},
address = {New York, NY, USA},
url = {https://doi.org/10.1145/3774904.3792931},
doi = {10.1145/3774904.3792931},
booktitle = {Proceedings of the ACM Web Conference 2026},
pages = {8649–8652},
numpages = {4},
location = {United Arab Emirates},
series = {WWW '26}
}

@inproceedings{hou2024blair,
    title = "Bridging Language and Items for Retrieval and Recommendation: Benchmarking {LLM}s as Semantic Encoders",
    author = "Hou, Yupeng  and
      Li, Jiacheng  and
      Fu, Xiangjun  and
      He, Zhankui  and
      Yan, An  and
      Chen, Xiusi  and
      McAuley, Julian",
    editor = "Liakata, Maria  and
      Moreira, Viviane P.  and
      Zhang, Jiajun  and
      Jurgens, David",
    booktitle = "Proceedings of the 64th Annual Meeting of the {A}ssociation for {C}omputational {L}inguistics (Volume 1: Long Papers)",
    month = jul,
    year = "2026",
    address = "San Diego, California, United States",
    publisher = "Association for Computational Linguistics",
    url = "https://aclanthology.org/2026.acl-long.147/",
    doi = "10.18653/v1/2026.acl-long.147",
    pages = "3251--3265",
    ISBN = "979-8-89176-390-6",
}

@article{zhang2026recthinker,
  title={RecThinker: An Agentic Framework for Tool-Augmented Reasoning in Recommendation},
  author={Zhang, Haobo and Zhu, Yutao and Mao, Kelong and Li, Tianhao and Dou, Zhicheng},
  journal={arXiv preprint arXiv:2603.09843},
  year={2026}
}

@inproceedings{zhao2025personalens,
    title = "{P}ersona{L}ens: A Benchmark for Personalization Evaluation in Conversational {AI} Assistants",
    author = "Zhao, Zheng  and
      Vania, Clara  and
      Kayal, Subhradeep  and
      Khan, Naila  and
      Cohen, Shay B.  and
      Yilmaz, Emine",
    editor = "Che, Wanxiang  and
      Nabende, Joyce  and
      Shutova, Ekaterina  and
      Pilehvar, Mohammad Taher",
    booktitle = "Findings of the Association for Computational Linguistics: ACL 2025",
    month = jul,
    year = "2025",
    address = "Vienna, Austria",
    publisher = "Association for Computational Linguistics",
    url = "https://aclanthology.org/2025.findings-acl.927/",
    doi = "10.18653/v1/2025.findings-acl.927",
    pages = "18023--18055",
    ISBN = "979-8-89176-256-5",
}

@article{yue2023llamarec,
  title={Llamarec: Two-stage recommendation using large language models for ranking},
  author={Yue, Zhenrui and Rabhi, Sara and Moreira, Gabriel de Souza Pereira and Wang, Dong and Oldridge, Even},
  journal={arXiv preprint arXiv:2311.02089},
  year={2023}
}

@inproceedings{hou2023llmrank,
  title={Large language models are zero-shot rankers for recommender systems},
  author={Hou, Yupeng and Zhang, Junjie and Lin, Zihan and Lu, Hongyu and Xie, Ruobing and McAuley, Julian and Zhao, Wayne Xin},
  booktitle={European conference on information retrieval},
  pages={364--381},
  year={2024},
  organization={Springer}
}

@inproceedings{ihemelandu2023candidate,
  title={Candidate set sampling for evaluating top-n recommendation},
  author={Ihemelandu, Ngozi and Ekstrand, Michael D},
  booktitle={2023 IEEE/WIC International Conference on Web Intelligence and Intelligent Agent Technology (WI-IAT)},
  pages={88--94},
  year={2023},
  organization={IEEE}
}

@inproceedings{krichene2020sampled,
author = {Krichene, Walid and Rendle, Steffen},
title = {On Sampled Metrics for Item Recommendation},
year = {2020},
isbn = {9781450379984},
publisher = {Association for Computing Machinery},
address = {New York, NY, USA},
url = {https://doi.org/10.1145/3394486.3403226},
doi = {10.1145/3394486.3403226},
booktitle = {Proceedings of the 26th ACM SIGKDD International Conference on Knowledge Discovery \& Data Mining},
pages = {1748–1757},
numpages = {10},
location = {Virtual Event, CA, USA},
series = {KDD '20}
}

@misc{ballou2025openplay, title={Open Play: A longitudinal dataset of multi-platform video game digital trace data and psychological measures}, url={osf.io/preprints/psyarxiv/nz96c_v1}, DOI={10.31234/osf.io/nz96c_v1}, publisher={PsyArXiv}, author={Ballou, Nick and Földes, Tamás A and Vuorre, Matti and Hakman, Thomas and Magnusson, Kristoffer and Przybylski, Andrew K}, year={2025}, month={Nov} }

@inproceedings{ju2025autocdsr,
author = {Ju, Clark Mingxuan and Neves, Leonardo and Kumar, Bhuvesh and Collins, Liam and Zhao, Tong and Qiu, Yuwei and Dou, Qing and Nizam, Sohail and Yang, Sen and Shah, Neil},
title = {Revisiting Self-attention for Cross-domain Sequential Recommendation},
year = {2025},
isbn = {9798400714542},
publisher = {Association for Computing Machinery},
address = {New York, NY, USA},
url = {https://doi.org/10.1145/3711896.3737108},
doi = {10.1145/3711896.3737108},
booktitle = {Proceedings of the 31st ACM SIGKDD Conference on Knowledge Discovery and Data Mining V.2},
pages = {1094–1105},
numpages = {12},
location = {Toronto ON, Canada},
series = {KDD '25}
}

@inproceedings{hou2022unisrec,
author = {Hou, Yupeng and Mu, Shanlei and Zhao, Wayne Xin and Li, Yaliang and Ding, Bolin and Wen, Ji-Rong},
title = {Towards Universal Sequence Representation Learning for Recommender Systems},
year = {2022},
isbn = {9781450393850},
publisher = {Association for Computing Machinery},
address = {New York, NY, USA},
url = {https://doi.org/10.1145/3534678.3539381},
doi = {10.1145/3534678.3539381},
pages = {585–593},
numpages = {9},
location = {Washington DC, USA},
series = {KDD '22}
}

@inproceedings{li2021recguru,
author = {Li, Chenglin and Zhao, Mingjun and Zhang, Huanming and Yu, Chenyun and Cheng, Lei and Shu, Guoqiang and Kong, BeiBei and Niu, Di},
title = {RecGURU: Adversarial Learning of Generalized User Representations for Cross-Domain Recommendation},
year = {2022},
isbn = {9781450391320},
publisher = {Association for Computing Machinery},
address = {New York, NY, USA},
url = {https://doi.org/10.1145/3488560.3498388},
doi = {10.1145/3488560.3498388},
booktitle = {Proceedings of the Fifteenth ACM International Conference on Web Search and Data Mining},
pages = {571–581},
numpages = {11},
location = {Virtual Event, AZ, USA},
series = {WSDM '22}
}

@inproceedings{
kim2026persona2web,
title={Persona2Web: Benchmarking Personalized Web Agents for Contextual Reasoning with User History},
author={Serin Kim and Sangam Lee and Dongha Lee},
booktitle={Forty-third International Conference on Machine Learning},
year={2026},
url={https://openreview.net/forum?id=qvvD9hgHoX}
}

@inproceedings{wang2026meagent,
    title = "Me-Agent: A Personalized Mobile Agent with Two-Level User Habit Learning for Enhanced Interaction",
    author = "Wang, Shuoxin  and
      Liu, Chang  and
      Loo, Gowen  and
      Zheng, Lifan  and
      Wei, Kaiwen  and
      Yan, Huanqian  and
      Zeng, Xinyi  and
      Zhang, Jingyuan  and
      Tian, Yu",
    editor = "Liakata, Maria  and
      Moreira, Viviane P.  and
      Zhang, Jiajun  and
      Jurgens, David",
    booktitle = "Findings of the {A}ssociation for {C}omputational {L}inguistics: {ACL} 2026",
    month = jul,
    year = "2026",
    address = "San Diego, California, United States",
    publisher = "Association for Computational Linguistics",
    url = "https://aclanthology.org/2026.findings-acl.1211/",
    doi = "10.18653/v1/2026.findings-acl.1211",
    pages = "24206--24222",
    ISBN = "979-8-89176-395-1",
}

@inproceedings{lyu2026personalalign,
    title = "{P}ersonal{A}lign: Hierarchical Implicit Intent Alignment for Personalized {GUI} Agent with Long-Term User-Centric Records",
    author = "Lyu, Yibo  and
      Chen, Gongwei  and
      Shao, Rui  and
      Guan, Weili  and
      Nie, Liqiang",
    editor = "Liakata, Maria  and
      Moreira, Viviane P.  and
      Zhang, Jiajun  and
      Jurgens, David",
    booktitle = "Proceedings of the 64th Annual Meeting of the {A}ssociation for {C}omputational {L}inguistics (Volume 1: Long Papers)",
    month = jul,
    year = "2026",
    address = "San Diego, California, United States",
    publisher = "Association for Computational Linguistics",
    url = "https://aclanthology.org/2026.acl-long.1669/",
    doi = "10.18653/v1/2026.acl-long.1669",
    pages = "36074--36089",
    ISBN = "979-8-89176-390-6",
}

@inproceedings{zhang2026personaagent,
    title = "{P}ersona{A}gent: Bridging Memory and Action for Personalized {LLM} Agents",
    author = "Zhang, Weizhi  and
      Zhang, Xinyang  and
      Zhang, Chenwei  and
      Yang, Liangwei  and
      Shang, Jingbo  and
      Wei, Zhepei  and
      Zou, Henry Peng  and
      Huang, Zijie  and
      Wang, Zhengyang  and
      Gao, Yifan  and
      Pan, Xiaoman  and
      Xiong, Lian  and
      Liu, Jingguo  and
      Yu, Philip S.  and
      Li, Xian",
    editor = "Liakata, Maria  and
      Moreira, Viviane P.  and
      Zhang, Jiajun  and
      Jurgens, David",
    booktitle = "Findings of the {A}ssociation for {C}omputational {L}inguistics: {ACL} 2026",
    month = jul,
    year = "2026",
    address = "San Diego, California, United States",
    publisher = "Association for Computational Linguistics",
    url = "https://aclanthology.org/2026.findings-acl.1315/",
    doi = "10.18653/v1/2026.findings-acl.1315",
    pages = "26421--26439",
    ISBN = "979-8-89176-395-1",
}

@article{xu2026personalizedagents,
  title={Toward personalized llm-powered agents: Foundations, evaluation, and future directions},
  author={Xu, Yue and Chen, Qian and Ma, Zizhan and Liu, Dongrui and Wang, Wenxuan and Wang, Xiting and Xiong, Li and Wang, Wenjie},
  journal={arXiv preprint arXiv:2602.22680},
  year={2026}
}

@article{tao2026sager,
  title={SAGER: Self-Evolving User Policy Skills for Recommendation Agent},
  author={Tao, Zhen and Lai, Riwei and Yu, Chenyun and Chen, Weixin and Chen, Li and Kong, Beibei and Cheng, Lei and Zhuo, Chengxiang and Li, Zang and Sun, Qingqiang},
  journal={arXiv preprint arXiv:2604.14972},
  year={2026}
}

@article{shen2026mars,
  title={Agentic Recommender System with Hierarchical Belief-State Memory},
  author={Shen, Xiang and Zhou, Yuhang and Wu, Yifan and Zhao, Zhuokai and Lin, Siyu and Huang, Lei and Zhong, Qianqian and Zhang, Lizhu and Zhang, Benyu and Fan, Xiangjun and others},
  journal={arXiv preprint arXiv:2605.14401},
  year={2026}
}

@inproceedings{chen2026memrec,
    title = "{M}em{R}ec: Collaborative Memory-Augmented Agentic Recommender System",
    author = "Chen, Weixin  and
      Zhao, Yuhan  and
      Huang, Jingyuan  and
      Ye, Zihe  and
      Ju, Mingxuan  and
      Zhao, Tong  and
      Shah, Neil  and
      Chen, Li  and
      Zhang, Yongfeng",
    editor = "Liakata, Maria  and
      Moreira, Viviane P.  and
      Zhang, Jiajun  and
      Jurgens, David",
    booktitle = "Proceedings of the 64th Annual Meeting of the {A}ssociation for {C}omputational {L}inguistics (Volume 1: Long Papers)",
    month = jul,
    year = "2026",
    address = "San Diego, California, United States",
    publisher = "Association for Computational Linguistics",
    url = "https://aclanthology.org/2026.acl-long.2061/",
    doi = "10.18653/v1/2026.acl-long.2061",
    pages = "44515--44544",
    ISBN = "979-8-89176-390-6",
}

@inproceedings{zhang2023agentcf,
author = {Zhang, Junjie and Hou, Yupeng and Xie, Ruobing and Sun, Wenqi and McAuley, Julian and Zhao, Wayne Xin and Lin, Leyu and Wen, Ji-Rong},
title = {AgentCF: Collaborative Learning with Autonomous Language Agents for Recommender Systems},
year = {2024},
isbn = {9798400701719},
publisher = {Association for Computing Machinery},
address = {New York, NY, USA},
url = {https://doi.org/10.1145/3589334.3645537},
doi = {10.1145/3589334.3645537},
booktitle = {Proceedings of the ACM Web Conference 2024},
pages = {3679–3689},
numpages = {11},
location = {Singapore, Singapore},
series = {WWW '24}
}

@inproceedings{wang2023recmind,
    title = "{R}ec{M}ind: Large Language Model Powered Agent For Recommendation",
    author = "Wang, Yancheng  and
      Jiang, Ziyan  and
      Chen, Zheng  and
      Yang, Fan  and
      Zhou, Yingxue  and
      Cho, Eunah  and
      Fan, Xing  and
      Lu, Yanbin  and
      Huang, Xiaojiang  and
      Yang, Yingzhen",
    editor = "Duh, Kevin  and
      Gomez, Helena  and
      Bethard, Steven",
    booktitle = "Findings of the Association for Computational Linguistics: NAACL 2024",
    month = jun,
    year = "2024",
    address = "Mexico City, Mexico",
    publisher = "Association for Computational Linguistics",
    url = "https://aclanthology.org/2024.findings-naacl.271/",
    doi = "10.18653/v1/2024.findings-naacl.271",
    pages = "4351--4364",
}

@article{huang2023interecagent,
author = {Huang, Xu and Lian, Jianxun and Lei, Yuxuan and Yao, Jing and Lian, Defu and Xie, Xing},
title = {Recommender AI Agent: Integrating Large Language Models for Interactive Recommendations},
year = {2025},
issue_date = {July 2025},
publisher = {Association for Computing Machinery},
address = {New York, NY, USA},
volume = {43},
number = {4},
issn = {1046-8188},
url = {https://doi.org/10.1145/3731446},
doi = {10.1145/3731446},
journal = {ACM Trans. Inf. Syst.},
month = jun,
articleno = {96},
numpages = {33}
}

@inproceedings{xu2025iagent,
    title = "i{A}gent: {LLM} Agent as a Shield between User and Recommender Systems",
    author = "Xu, Wujiang  and
      Shi, Yunxiao  and
      Liang, Zujie  and
      Ning, Xuying  and
      Mei, Kai  and
      Wang, Kun  and
      Zhu, Xi  and
      Xu, Min  and
      Zhang, Yongfeng",
    editor = "Che, Wanxiang  and
      Nabende, Joyce  and
      Shutova, Ekaterina  and
      Pilehvar, Mohammad Taher",
    booktitle = "Findings of the Association for Computational Linguistics: ACL 2025",
    month = jul,
    year = "2025",
    address = "Vienna, Austria",
    publisher = "Association for Computational Linguistics",
    url = "https://aclanthology.org/2025.findings-acl.928/",
    doi = "10.18653/v1/2025.findings-acl.928",
    pages = "18056--18084",
    ISBN = "979-8-89176-256-5",
}

@inproceedings{covington2016youtube,
author = {Covington, Paul and Adams, Jay and Sargin, Emre},
title = {Deep Neural Networks for YouTube Recommendations},
year = {2016},
isbn = {9781450340359},
publisher = {Association for Computing Machinery},
address = {New York, NY, USA},
url = {https://doi.org/10.1145/2959100.2959190},
doi = {10.1145/2959100.2959190},
booktitle = {Proceedings of the 10th ACM Conference on Recommender Systems},
pages = {191–198},
numpages = {8},
location = {Boston, Massachusetts, USA},
series = {RecSys '16}
}

@article{liu2025onerecthink,
  title={Onerec-think: In-text reasoning for generative recommendation},
  author={Liu, Zhanyu and Wang, Shiyao and Wang, Xingmei and Zhang, Rongzhou and Deng, Jiaxin and Bao, Honghui and Zhang, Jinghao and Li, Wuchao and Zheng, Pengfei and Wu, Xiangyu and others},
  journal={arXiv preprint arXiv:2510.11639},
  year={2025}
}

@inproceedings{kwon2023efficient,
  title={Efficient Memory Management for Large Language Model Serving with PagedAttention},
  author={Woosuk Kwon and Zhuohan Li and Siyuan Zhuang and Ying Sheng and Lianmin Zheng and Cody Hao Yu and Joseph E. Gonzalez and Hao Zhang and Ion Stoica},
  booktitle={Proceedings of the ACM SIGOPS 29th Symposium on Operating Systems Principles},
  year={2023}
}

@article{sheng2024hybridflow,
  title   = {HybridFlow: A Flexible and Efficient RLHF Framework},
  author  = {Guangming Sheng and Chi Zhang and Zilingfeng Ye and Xibin Wu and Wang Zhang and Ru Zhang and Yanghua Peng and Haibin Lin and Chuan Wu},
  year    = {2024},
  journal = {arXiv preprint arXiv: 2409.19256}
}

@article{shao2024deepseekmath,
  title={Deepseekmath: Pushing the limits of mathematical reasoning in open language models},
  author={Shao, Zhihong and Wang, Peiyi and Zhu, Qihao and Xu, Runxin and Song, Junxiao and Bi, Xiao and Zhang, Haowei and Zhang, Mingchuan and Li, YK and Wu, Yang and others},
  journal={arXiv preprint arXiv:2402.03300},
  year={2024}
}

@inproceedings{
zhu2025rankgrpo,
title={Rank-{GRPO}: Training {LLM}-based Conversational Recommender Systems with Reinforcement Learning},
author={Yaochen Zhu and Harald Steck and Dawen Liang and Yinhan He and Vito Claudio Ostuni and Jundong Li and Nathan Kallus},
booktitle={The Fourteenth International Conference on Learning Representations},
year={2026},
url={https://openreview.net/forum?id=Xgw2D9cALS}
}

@inproceedings{huang2026rerec,
    title = "{R}e{R}ec: Reasoning-Augmented {LLM}-based Recommendation Assistant via Reinforcement Fine-tuning",
    author = "Huang, Jiani  and
      Wang, Shijie  and
      Ning, Liang-Bo  and
      Fan, Wenqi  and
      Qing, Li",
    editor = "Liakata, Maria  and
      Moreira, Viviane P.  and
      Zhang, Jiajun  and
      Jurgens, David",
    booktitle = "Proceedings of the 64th Annual Meeting of the {A}ssociation for {C}omputational {L}inguistics (Volume 1: Long Papers)",
    month = jul,
    year = "2026",
    address = "San Diego, California, United States",
    publisher = "Association for Computational Linguistics",
    url = "https://aclanthology.org/2026.acl-long.964/",
    doi = "10.18653/v1/2026.acl-long.964",
    pages = "21040--21055",
    ISBN = "979-8-89176-390-6",
}

@article{
lin2025recr1,
title={Rec-R1: Bridging Generative Large Language Models and User-Centric Recommendation Systems via Reinforcement Learning},
author={Jiacheng Lin and Tian Wang and Kun Qian},
journal={Transactions on Machine Learning Research},
issn={2835-8856},
year={2025},
url={https://openreview.net/forum?id=YBRU9MV2vE},
note={}
}

@inproceedings{kang2018sasrec,
  title={Self-attentive sequential recommendation},
  author={Kang, Wang-Cheng and McAuley, Julian},
  booktitle={2018 IEEE international conference on data mining (ICDM)},
  pages={197--206},
  year={2018},
  organization={IEEE}
}

@article{narasimhan2026taurec,
  title={$\tau$-Rec: A Verifiable Benchmark for Agentic Recommender Systems},
  author={Narasimhan, Bharath Sivaram and Narasimhan, Karthik R},
  journal={arXiv preprint arXiv:2606.10156},
  year={2026}
}

@inproceedings{
shang2025agentrecbench,
title={AgentRecBench: Benchmarking {LLM} Agent-based Personalized Recommender Systems},
author={Yu Shang and Peijie Liu and Yuwei Yan and Zijing Wu and Leheng Sheng and Yuanqing Yu and Chumeng Jiang and An Zhang and Fengli Xu and Yu Wang and Min Zhang and Yong Li},
booktitle={The Thirty-ninth Annual Conference on Neural Information Processing Systems Datasets and Benchmarks Track},
year={2026},
url={https://openreview.net/forum?id=fm77rDf9JS}
}

@article{wu2026clawrec,
  title={ClawRec: A Claw-Native Recommender System},
  author={Wu, Chenghao and Ou, Kesha and Wang, Xiaolei and Zheng, Bowen and Li, Bingqian and Liu, Enze and Zhao, Wayne Xin and Li, Weitao and Zhang, Long and Chen, Sheng and others},
  journal={arXiv preprint arXiv:2607.23779},
  year={2026}
}

@article{lin2026agenticroadmap,
  title={Autonomous Information Seeking: A Roadmap for Agentic Recommender Systems},
  author={Lin, Xinyu and Deldjoo, Yashar and Dai, Sunhao and Bao, Honghui and Ye, Xiaopeng and Nazary, Fatemeh and Wang, Wenjie and Di Noia, Tommaso and Xu, Jun and Chua, Tat-Seng},
  journal={arXiv preprint arXiv:2607.04433},
  year={2026}
}

@article{huang2025agenticrec,
  title={Towards agentic recommender systems in the era of multimodal large language models},
  author={Huang, Chengkai and Wu, Junda and Xia, Yu and Yu, Zixu and Wang, Ruhan and Yu, Tong and Zhang, Ruiyi and Rossi, Ryan A and Kveton, Branislav and Zhou, Dongruo and others},
  journal={arXiv preprint arXiv:2503.16734},
  year={2025}
}

@article{sun2026autonomousagents,
  title={A Position Paper on Recommender Systems in the Era of Autonomous Agents},
  author={Sun, Aixin},
  journal={arXiv preprint arXiv:2607.24822},
  year={2026}
}

@article{zhang2026usercentric,
  title={The Next Paradigm Is User-Centric Agent, Not Platform-Centric Service},
  author={Zhang, Luankang and Lv, Hang and Pan, Qiushi and Wang, Kefen and Huang, Yonghao and Miao, Xinrui and Xu, Yin and Guo, Wei and Liu, Yong and Wang, Hao and others},
  journal={arXiv:2602.15682},
  year={2026}
}

\clearpage
\newpage
\beginappendix

This appendix provides additional details for benchmark reproducibility and theoretical analysis. Appendix \ref{app:benchmark} includes the futher construction details of {MediateRec}, target-platform LLM reranking prompt, and personal-agent mediation prompt. Appendix \ref{app:pamo} provides further implementation details for PAMO, derivations of its NDCG decomposition and value-constrained allocation, complete proofs of the theoretical results, and additional training details.

\section{{MediateRec} Benchmark Details}
\label{app:benchmark}

This appendix specifies the source processing, temporal episode construction, candidate sampling, platform generation, and personal-agent inputs defered from Section~\ref{sec:benchmark}.

\subsection{Amazon Proxy Construction}

\subsubsection{\textbf{Source and Episodes}}
We use the rating-only 0-core population of Amazon Reviews 2023 keyed by parent ASIN \citep{hou2024blair}. Products without titles are removed, metadata are deduplicated by parent ASIN, and repeated user--item interactions retain the earliest timestamp. A user is eligible after interacting in at least two Amazon categories. For each retained episode, the user's final interaction on the target platform is held out as $Y$; earlier target-platform interactions form within-platform history, and earlier interactions in other Amazon categories form cross-platform history. Both histories must be nonempty and must strictly precede the target timestamp. Within-platform records contain the item title, category, and star rating; cross-platform records additionally identify the source Amazon category. Across the construction pools, 33 non-target categories appear in cross-platform histories.

Each user contributes one episode in exactly one target platform. Eligible users are shuffled with seed 7 and processed in the fixed order Movie, Toy, Grocery, and Beauty, with a global used-user set assigning a user eligible for several targets to the first one in this order. This guarantees disjoint users across target platforms and splits, although the fixed order can favor earlier target platforms. Split sizes and evaluation roles are reported in Table~\ref{tab:benchmark_stats}.

\subsubsection{\textbf{Candidate Sets}}
Each episode contains $Y$ and 49 distinct negatives from the titled catalog of the target platform. Negatives are sampled without replacement in proportion to their target-category interaction counts over the full population. The target and every item in the user's histories are excluded. The catalog and popularity counts are static rather than restricted to items observed before the target timestamp. The 50 candidates are randomly permuted with seed 7 before platform construction.

\subsection{OpenPlay Construction}

\subsubsection{\textbf{Source and cohort}}
We use OpenPlay \citep{ballou2025openplay}, which contains pseudonymized person identifier links Steam, Nintendo, and Xbox records for the same person. Steam and Nintendo records expose game titles, whereas Xbox records expose genres but not resolved game titles. We obtain Steam genres from IGDB metadata and omit mobile records because they contain aggregate activity without game identity. Starting from 732 users with Steam activity and at least one Nintendo or Xbox record, the target and pre-target history requirements described below yield the 645-user external-test cohort.

\subsubsection{\textbf{Temporal Episodes and Histories}}
For each user, $Y$ is the last-adopted Steam title among games with at least 30 minutes of total playtime. Adoption time is the title's first-play timestamp, which also defines the episode split time $T$. Within-platform history contains distinct Steam titles first played strictly before $T$, while cross-platform history contains Nintendo and Xbox records strictly before $T$. Each episode requires at least one record of each type.

Within-platform history is rendered most-recent-first and capped at 30 Steam titles; cross-platform history is capped at 50 records. Steam records contain a title, primary genre, and a five-level engagement value based on cumulative playtime observed strictly before $T$, using boundaries of 0.5, 2, 10, and 50 hours. Nintendo and Xbox engagement is computed analogously from pre-$T$ activity. Engagement is rendered in a dedicated field rather than as an explicit rating. Nintendo records retain titles, Xbox records retain genres, and every cross-platform record identifies its source service.

\subsubsection{\textbf{Candidate Sets}}
Each episode contains $Y$ and 49 Steam negatives sampled without replacement in proportion to title popularity, using exponent $1.0$ and seed 7. Previously played Steam titles are excluded. Candidates are randomly permuted before platform construction using seed $7+\texttt{index}$ for each episode.

\subsection{Target Platform Construction}\label{app:rerank}

For each target platform, we first train a SASRec model on chronological
within-platform interactions \citep{kang2018sasrec}. The models use
64 dimensional item and positional embeddings, maximum sequence length 50,
two self-attention blocks, one attention head, dropout $0.5$, and sampled
cross-entropy with 256 uniformly sampled negatives. Training uses Adam with
learning rate $10^{-3}$ and batch size 256, with early stopping on validation
HR@10.

For each benchmark episode, SASRec ranks the fixed 50-item candidate set using
within-platform information only. Claude Sonnet 4.6 then reranks this
SASRec-ordered list at temperature 0 using the user's within-platform history.
Cross-platform history is never exposed during platform construction. The
resulting ranking is frozen for all personal-agent training and evaluation,
and candidates are relabeled by this final order so that \texttt{C01} denotes
the platform's highest-ranked candidate.

\subsubsection{\textbf{Platform LLM Reranking Prompt}}

We use the same platform-reranking instruction across datasets, with
dataset-specific history and item fields substituted for the placeholders
below.

\begin{tcolorbox}[
    colback=gray!3,
    colframe=gray!80,
    boxrule=0.5pt,
    arc=2pt,
    left=5pt,
    right=5pt,
    top=4pt,
    bottom=4pt,
    title=\textbf{System}
]
\small
You are the \{Domain\} platform's recommendation ranker. You are given a list
of candidate items that an upstream collaborative-filtering model has already
ranked for this user (best guess first), together with the user's recent
activity on this platform. Your job is to re-rank the candidates into the
order most likely to match what this user will engage with next, using the
platform's ranking as a strong prior and the user's on-platform history as
supporting evidence. Return a best-first ordering of the candidates.
\end{tcolorbox}

\begin{tcolorbox}[
    colback=gray!3,
    colframe=gray!80,
    boxrule=0.5pt,
    arc=2pt,
    left=5pt,
    right=5pt,
    top=4pt,
    bottom=4pt,
    title=\textbf{User}
]
\small
Re-rank the \{Domain\} platform's candidate recommendations for this user.

\medskip
The user's recent activity on the \{Domain\} platform (most recent first):

\texttt{- Title [cat] (rated X)}

\medskip
The current ranked candidates (the model's best guess first):

\texttt{C01: Title [cat]} \ldots{} \texttt{C50: \ldots}

\medskip
Order all candidates from most to least likely for this user. On the LAST line
output exactly:

\texttt{FINAL: ["C..", ... best first]}
\end{tcolorbox}

For Open Play, product categories and ratings are replaced by game genres and
the corresponding engagement field. The platform prompt never contains
Nintendo or Xbox history.

\subsection{Personal-Agent Interface}\label{app:interface}

The personal agent receives the 50 candidates in frozen platform order, up to
30 within-platform records, and up to 50 cross-platform records. Candidate
cards contain an identifier, title, and public category or genre. Amazon
histories contain titles, categories, and star ratings, with the source
category attached to cross-platform records. Open Play histories contain
Steam titles and engagement values, Nintendo titles, or Xbox genres, together
with their source services.

The agent returns ten distinct candidate identifiers in best-first order after
a \texttt{FINAL:} marker. It cannot introduce items outside the candidate set
and does not observe platform scores, population interactions, or internal
model states. Agent-training examples are constructed only from Movie and Toy;
the held-out target identifier is used for reward computation but never
appears in the model prompt.

\subsubsection{\textbf{Default Mediation Prompt}}

All personal-agent conditions use the same mediation instruction. The prompt
treats the platform ranking as meaningful evidence while asking the agent to
intervene only when the user's cross-platform history provides sufficiently
specific support.

\begin{tcolorbox}[
    colback=gray!3,
    colframe=gray!80,
    boxrule=0.5pt,
    arc=2pt,
    left=5pt,
    right=5pt,
    top=4pt,
    bottom=4pt,
    title=\textbf{System}
]
\small
You are this user's personal recommendation agent, authorized to access their
historical interactions across platforms, acting in the user's interest. The
platform's recommender provides a ranked candidate list; its ranking already
encodes collaborative signal from many similar users that you cannot directly
see, so a high rank is real evidence in its own right. Your job is to mediate:
decide where the platform's ranking already serves this user and should be
KEPT, and where this user's private cross-platform history gives you specific,
on-point evidence to move a lower-ranked item up. Deferring to a strong
platform ranking is often the right call---intervene decisively only when your
private evidence is specific, not generic or loosely related. Reason about who
the user is, judge where your evidence is strong versus weak, then give your
final selection.
\end{tcolorbox}

\begin{tcolorbox}[
    colback=gray!3,
    colframe=gray!80,
    boxrule=0.5pt,
    arc=2pt,
    left=5pt,
    right=5pt,
    top=4pt,
    bottom=4pt,
    title=\textbf{User}
]
\small
Mediate the platform's \{Domain\} recommendations for your user: choose the top
items to present.

\medskip
Your user's history on the \{Domain\} platform (most recent first):

\texttt{- \{within items: Title [cat] (rated X)\}}

\medskip
Your user's interactions on other platforms (use them to infer their tastes,
lifestyle, and situation), most recent first:

\texttt{- (\{SourceCat\}) Title [cat] (rated X)}

\medskip
The \{Domain\} platform recommender's current ranked candidates (its best guess
first):

\texttt{C01: Title [cat]} \ldots{} \texttt{C50: \ldots}

\medskip
Consider who this user is, and where your private evidence is specific enough
to override the platform's ranking versus where you should keep it. Keep the
platform's order where your evidence is weak, generic, or only loosely related;
move items up only with specific support. Then on the LAST line output exactly:

\texttt{FINAL: ["C..", ... 10 ids best first]}
\end{tcolorbox}

For Open Play, the same prompt structure is used with Steam titles, genres, and
engagement values for within-platform history; Nintendo titles and Xbox genres
for cross-platform history; and game genres in place of Amazon product
categories.

\subsection{Validation and Scope}

Every benchmark episode contains 50 distinct candidates and one target. All history records precede the target event, and sampled negatives exclude the target and the user's prior items. Because the target is inserted into every candidate set, {MediateRec} evaluates candidate-conditioned ranking and mediation rather than end-to-end retrieval.

The Amazon construction uses static catalogs and full-population popularity counts that are not restricted to items available before each target timestamp. Open Play provides focused external validation with Steam as the only target platform, Xbox evidence available at the genre level, and mobile activity omitted because game identities are unavailable. The benchmark release includes split assignments, fixed candidate sets, frozen platform rankings, normalized title mappings, prompt templates, and parser code.

\section{PAMO Derivations, Proofs, and  Details}
\label{app:pamo}

\subsection{NDCG Decomposition and Group-Relative Consistency}
\label{app:pamo_ndcg}

Suppose the relevant item appears at rank $q\leq K$. By Eq.~\eqref{eq:cutoff_outcomes}, $z_{g,k}=1$ exactly for $k\geq q$. Therefore,
\begin{equation}
\sum_{k=1}^{K}
\alpha_k z_{g,k}
=
\sum_{k=q}^{K}
(\ell_k-\ell_{k+1})
=
\ell_q.
\label{eq:app_ndcg_telescoping}
\end{equation}
If the relevant item is absent from the slate, every $z_{g,k}$ is zero. Hence, Eq.~\eqref{eq:ndcg_decomposition} exactly represents the NDCG utility in Eq.~\eqref{eq:ranking_utilities}.

Fix an active cutoff $k$. At $\eta=0$, the unique solution of Eq.~\eqref{eq:pamo_allocation} is the uniform distribution $u_k$, which satisfies the value floor with equality. If the platform misses at cutoff $k$, $\mathcal D_k$ contains the $Gp_k$ successful rollouts, and each receives
\begin{equation}
\frac{m_k}{Gp_k}
=
1-p_k.
\label{eq:app_uniform_rescue}
\end{equation}
If the platform succeeds, $\mathcal D_k$ contains the $G(1-p_k)$ failures, and each receives
\begin{equation}
-\frac{m_k}{G(1-p_k)}
=
-p_k.
\label{eq:app_uniform_harm}
\end{equation}
Thus, at every active cutoff,
\begin{equation}
A_{g,k}^{\mathrm{PAMO}}
=
z_{g,k}-p_k
\qquad
\text{when }\eta=0.
\label{eq:app_uniform_cutoff_advantage}
\end{equation}
At an inactive cutoff, $p_k\in\{0,1\}$, so $m_k=0$ and both the standard
group-relative advantage and the PAMO advantage are identically zero.

Since $\bar r=G^{-1}\sum_g r_g=\sum_k\alpha_kp_k$, recombining the cutoff-level advantages gives
\begin{equation}
\sum_{k=1}^{K}
\alpha_k
(z_{g,k}-p_k)
=
r_g-\bar r.
\label{eq:app_centered_ndcg}
\end{equation}
The uniform allocation therefore exactly matches centered NDCG supervision.

\subsection{Dual Form of the PAMO Allocation}
\label{app:pamo_dual}

For one active cutoff, the Lagrangian of Eq.~\eqref{eq:pamo_allocation} is
\begin{equation}
\begin{aligned}
\mathcal J(w,\lambda,\nu)
={}&
\eta
\sum_{g\in\mathcal D_k}
w_gc_g
-
\operatorname{KL}(w\Vert u_k)
\\
&+
\lambda
\left(
\sum_{g\in\mathcal D_k}
w_gv_{g,k}
-
\bar v_k
\right)
+
\nu
\left(
\sum_{g\in\mathcal D_k}
w_g-1
\right),
\end{aligned}
\label{eq:app_pamo_lagrangian}
\end{equation}
where $\lambda\geq0$. Stationarity with respect to $w_g$ gives
\begin{equation}
\log
\frac{w_g}{u_{g,k}}
=
\eta c_g+\lambda v_{g,k}+\mathrm{const}.
\label{eq:app_pamo_stationarity}
\end{equation}
Since $u_k$ is uniform, normalization yields Eq.~\eqref{eq:pamo_weights}.

Complementary slackness gives
\begin{equation}
\lambda_k
\left(
\mathbb E_{w_k^\star}[v_k]
-
\mathbb E_{u_k}[v_k]
\right)
=
0.
\label{eq:app_complementary_slackness}
\end{equation}
Thus, $\lambda_k=0$ whenever the support-only allocation satisfies the value floor. Otherwise, the constraint is active and holds with equality.

For fixed $\eta$, the weighted direction-aligned NDCG magnitude is nondecreasing in $\lambda_k$:
\begin{equation}
\frac{d}{d\lambda_k}
\mathbb E_{w_k^\star}[v_k]
=
\operatorname{Var}_{w_k^\star}(v_k)
\geq0.
\label{eq:app_dual_monotonicity}
\end{equation}
The derivative is strictly positive whenever the values
$\{v_{g,k}:g\in\mathcal D_k\}$ are not all equal. Hence, if the
unconstrained support tilt violates the value floor, the active dual
coefficient is unique and can be found by one-dimensional root finding.
If all $v_{g,k}$ are equal, the value floor is vacuous and we choose
$\lambda_k=0$.

\subsection{Proof of Proposition~\ref{prop:pamo_preservation}}
\label{app:proof_preservation}

Advantage preservation follows from the normalization of $w_k^\star$. If the platform misses, the rescue advantages satisfy
\begin{equation}
\sum_{g:z_{g,k}=1}
A_{g,k}^{\mathrm{PAMO}}
=
m_k
\sum_{g\in\mathcal D_k}
w_{g,k}^\star
=
m_k.
\label{eq:app_preserve_miss}
\end{equation}
The failed rollouts retain advantage $-p_k$ and sum to $-m_k$. If the platform succeeds, the successful rollouts retain advantage $1-p_k$ and sum to $m_k$, while
\begin{equation}
\sum_{g:z_{g,k}=0}
A_{g,k}^{\mathrm{PAMO}}
=
-m_k
\sum_{g\in\mathcal D_k}
w_{g,k}^\star
=
-m_k.
\label{eq:app_preserve_hit}
\end{equation}
This proves Eq.~\eqref{eq:advantage_preservation}.

The first inequality in Eq.~\eqref{eq:value_attribution_guarantees} is the feasibility requirement in Eq.~\eqref{eq:pamo_allocation}. To prove the second, note that $u_k$ is feasible. Optimality of $w_k^\star$ therefore gives
\begin{equation}
\eta
\mathbb E_{w_k^\star}[c]
-
\operatorname{KL}(w_k^\star\Vert u_k)
\geq
\eta
\mathbb E_{u_k}[c].
\label{eq:app_attribution_optimality}
\end{equation}
For $\eta>0$,
\begin{equation}
\mathbb E_{w_k^\star}[c]
\geq
\mathbb E_{u_k}[c]
+
\frac{1}{\eta}
\operatorname{KL}(w_k^\star\Vert u_k)
\geq
\mathbb E_{u_k}[c].
\label{eq:app_attribution_improvement}
\end{equation}
At $\eta=0$, $w_k^\star=u_k$, so the inequality holds with equality.

\subsection{Proof of Theorem~\ref{thm:pamo_local}}
\label{app:proof_local}

Let $n_k=|\mathcal D_k|$, and let $w_k(\eta,\lambda)$ denote the
normalized exponential weights
\[
w_{g,k}(\eta,\lambda)
=
\frac{
\exp\!\left(\eta c_g+\lambda v_{g,k}\right)
}{
\sum_{j\in\mathcal D_k}
\exp\!\left(\eta c_j+\lambda v_{j,k}\right)
}.
\]
At $\eta=0$ and $\lambda=0$, these weights equal the uniform distribution
$u_k$.

If $\widetilde v_k=0$, then $v_{g,k}$ is constant over
$\mathcal D_k$, so the value floor is satisfied with equality by every
allocation. We may therefore choose $\lambda_k(\eta)=0$. The softmax
expansion below then holds with $\mu_k=0$.

Now suppose $\widetilde v_k\neq0$. Define
\begin{equation}
F_k(\eta,\lambda)
=
\mathbb E_{w_k(\eta,\lambda)}[v_k]
-
\mathbb E_{u_k}[v_k].
\label{eq:app_value_boundary}
\end{equation}
At $(\eta,\lambda)=(0,0)$,
\begin{equation}
F_k(0,0)=0,
\qquad
\frac{\partial F_k}{\partial\eta}(0,0)
=
\operatorname{Cov}_{u_k}(c,v_k)
=
\frac{1}{n_k}
\left\langle
\widetilde c_k,\widetilde v_k
\right\rangle,
\label{eq:app_value_eta_derivative}
\end{equation}
and
\begin{equation}
\frac{\partial F_k}{\partial\lambda}(0,0)
=
\operatorname{Var}_{u_k}(v_k)
=
\frac{1}{n_k}
\left\|
\widetilde v_k
\right\|_2^2
>
0.
\label{eq:app_value_lambda_derivative}
\end{equation}
The implicit function theorem therefore gives a differentiable boundary
$\lambda_k^{\mathrm{bd}}(\eta)$ near $\eta=0$ such that
\[
F_k\!\left(\eta,\lambda_k^{\mathrm{bd}}(\eta)\right)=0,
\qquad
\lambda_k^{\mathrm{bd}}(0)=0,
\]
with
\begin{equation}
\left.
\frac{d\lambda_k^{\mathrm{bd}}}{d\eta}
\right|_{\eta=0}
=
-
\frac{
\left\langle
\widetilde c_k,\widetilde v_k
\right\rangle
}{
\left\|
\widetilde v_k
\right\|_2^2
}.
\label{eq:app_boundary_dual_derivative}
\end{equation}

The KKT conditions select $\lambda_k(\eta)=0$ when the unconstrained
support tilt is feasible, i.e., when $F_k(\eta,0)\geq0$, and otherwise
select the active boundary
$\lambda_k(\eta)=\lambda_k^{\mathrm{bd}}(\eta)$. Consequently,
\begin{equation}
\mu_k
\equiv
\lim_{\eta\downarrow0}
\frac{\lambda_k(\eta)}{\eta}
=
\max\left\{
0,
-
\frac{
\left\langle
\widetilde c_k,\widetilde v_k
\right\rangle
}{
\left\|
\widetilde v_k
\right\|_2^2
}
\right\}.
\label{eq:app_dual_first_order}
\end{equation}
This expression also covers the boundary case
$\langle\widetilde c_k,\widetilde v_k\rangle=0$, for which
$\lambda_k(\eta)=o(\eta)$.

Expanding the softmax weights around $(\eta,\lambda)=(0,0)$ gives
\begin{equation}
w_{g,k}^\star
=
\frac{1}{n_k}
+
\frac{\eta}{n_k}
\left(
\widetilde c_{g,k}
+
\mu_k\widetilde v_{g,k}
\right)
+
o(\eta).
\label{eq:app_weight_expansion}
\end{equation}
Define
\begin{equation}
q_k
=
\widetilde c_k+\mu_k\widetilde v_k
=
\begin{cases}
\widetilde c_k,
&
\left\langle
\widetilde c_k,\widetilde v_k
\right\rangle\geq0
\ \text{or }\widetilde v_k=0,
\\[6pt]
\displaystyle
\widetilde c_k
-
\frac{
\left\langle
\widetilde c_k,\widetilde v_k
\right\rangle
}{
\left\|
\widetilde v_k
\right\|_2^2
}
\widetilde v_k,
&
\left\langle
\widetilde c_k,\widetilde v_k
\right\rangle<0.
\end{cases}
\label{eq:app_projected_direction}
\end{equation}
Because both centered vectors lie in the simplex tangent space,
$q_k$ is exactly the Euclidean projection of $\widetilde c_k$ onto
\[
\mathcal K_k
=
\left\{
q:
\mathbf 1^\top q=0,\;
\langle q,\widetilde v_k\rangle\geq0
\right\}.
\]
Equation~\eqref{eq:app_weight_expansion} therefore yields
\[
\left.
\frac{d w_k^\star}{d\eta}
\right|_{\eta=0^+}
=
\frac{1}{n_k}q_k,
\]
which proves Eq.~\eqref{eq:pamo_weight_first_order}. The corresponding
cutoff-advantage derivative follows from
$A_{g,k}^{\mathrm{PAMO}}=s_km_kw_{g,k}^\star$:
\begin{equation}
\left.
\frac{\partial A_{g,k}^{\mathrm{PAMO}}}{\partial\eta}
\right|_{\eta=0^+}
=
s_k
\frac{m_k}{n_k}
q_{g,k}.
\label{eq:pamo_first_order}
\end{equation}

It remains to prove directional optimality. Consider any infinitesimal
weight reallocation $h$ satisfying
\begin{equation}
\mathbf 1^\top h=0,
\qquad
\langle h,\widetilde v_k\rangle\geq0,
\qquad
\|h\|_2\leq1.
\label{eq:app_admissible_direction}
\end{equation}
The corresponding first-order increase in expected personal mediation
support is
\begin{equation}
\left\langle
h,\widetilde c_k
\right\rangle.
\label{eq:app_first_order_attribution}
\end{equation}

If
$\langle\widetilde c_k,\widetilde v_k\rangle\geq0$, then
$q_k=\widetilde c_k$ is feasible, and
$q_k/\|q_k\|_2$ maximizes
Eq.~\eqref{eq:app_first_order_attribution} by the Cauchy--Schwarz
inequality whenever $q_k\neq0$.

If
$\langle\widetilde c_k,\widetilde v_k\rangle<0$, write
\begin{equation}
\widetilde c_k
=
q_k
+
a_k\widetilde v_k,
\qquad
a_k
=
\frac{
\langle\widetilde c_k,\widetilde v_k\rangle
}{
\|\widetilde v_k\|_2^2
}
<0.
\label{eq:app_projection_decomposition}
\end{equation}
For every admissible $h$,
\begin{equation}
\left\langle
h,\widetilde c_k
\right\rangle
=
\langle h,q_k\rangle
+
a_k
\langle h,\widetilde v_k\rangle
\leq
\langle h,q_k\rangle
\leq
\|q_k\|_2.
\label{eq:app_projection_bound}
\end{equation}
Since
$\langle q_k,\widetilde v_k\rangle=0$, equality is attained by
$h=q_k/\|q_k\|_2$ whenever $q_k\neq0$. Thus,
$q_k/\|q_k\|_2$ maximizes the admissible first-order support gain. If
$q_k=0$, no admissible reallocation yields a positive first-order gain.

\subsection{Training Objective and Cost}
\label{app:pamo_training}

We use $A_g^{\mathrm{PAMO}}$ as the response-level advantage in the same
clipped token-level GRPO objective as the matched baseline. For generated
token $y_{g,t}$,
\begin{equation}
\rho_{g,t}(\theta)
=
\frac{
\pi_\theta(y_{g,t}\mid B,H,y_{g,<t})
}{
\pi_{\bar\theta}(y_{g,t}\mid B,H,y_{g,<t})
},
\label{eq:pamo_policy_ratio}
\end{equation}
and
\begin{equation}
\mathcal L_{\mathrm{PAMO}}
=
-\mathbb E_{g,t}\Big[
\min\Big\{
\rho_{g,t}(\theta)A_g^{\mathrm{PAMO}}, 
\operatorname{clip}\!\left(
\rho_{g,t}(\theta),1-\epsilon,1+\epsilon
\right)
A_g^{\mathrm{PAMO}}
\Big\}
\Big].
\label{eq:pamo_objective}
\end{equation}
The same scalar advantage is applied to all generated tokens in response
$g$; padding is masked, and the support scores and allocation weights are
detached from gradient computation. The full-history rollout
log-probabilities form the denominator in Eq.~\eqref{eq:pamo_policy_ratio}.

PAMO adds one teacher-forced scoring pass under the within-only input to
compute personal mediation support, but does not generate additional
rollouts. The constrained allocation is solved independently at each active
cutoff over at most $G$ responses, with a one-dimensional search for the dual
coefficient when the value floor is active. PAMO therefore leaves the rollout
budget unchanged and adds no inference-time component.

\end{document}